# Synthetic Consumer Insight Generation with Large Language Models

**Stephen L. France (Corresponding Author)**

Mississippi State University

Mailstop 9582, Mississippi State, MS 39762

Phone: +1 662-325-1630, Fax: +1-662-325-7012

sfrance@business.msstate.edu

ORCID iD: 0000-0002-7226-2750

**Pia A. Albinsson**

Beroth Professor of Marketing

Appalachian State University

ASU Box 32090

Boone, NC 28608-2090

albinssonpa@appstate.edu

ORCID iD:0000-0002-8591-5191

**Synthetic Consumer Insight Generation with Large Language Models**

## ABSTRACT

Modern data-driven marketing relies on large amounts of consumer data, yet collecting such data can be costly, time-consuming, and difficult to scale. This research examines whether large language models (LLMs) can be used to generate synthetic consumer data for projective techniques, a set of methods designed to elicit consumer associations, emotions, wants, and needs. We test LLM-generated responses across multiple projective tasks, LLMs, prompting strategies, and temperature settings, and compare them with human responses from a primary research study on perceptions of city tourism destinations. Human and LLM responses were analyzed using linguistic measures, diversity and concentration metrics, topic models, and top-term analyses. The results show substantial overlap between human and LLM responses in broad topics and associations, but also important differences in style, linguistic structure, and the way diversity is generated. Recommendations are given on how to best utilize LLMs for generating synthetic consumer data, how model and prompt choices shape response quality, and on recognizing the limitations of LLM synthetic consumer data generation.

## INTRODUCTION

Generative AI (artificial intelligence) using LLMs (large language models) is increasingly used to generate new datasets and augment existing datasets in applications as diverse as generating training data to improve machine learning performance (e.g., Li et al., 2023; Tang et al., 2023), generating health data for clinical discovery and disease prediction (Kang et al., 2024; Smolyak et al., 2024), and generating synthetic personality personas that can be used to simulate different types of users, customers, or decision makers in applied research tasks (Ge et al., 2024). These data "generation" capabilities can easily be applied to consumer behavior settings, where marketers require consumer data to help understand consumers' preferences and motivations (e.g., Basu et al., 2023), but where gathering data can be expensive and time consuming (Market Research Society, 2024). A key aim of synthetic data generation is to produce output that has similar data, including key data points and word distributions, to human responses (Rossi et al., 2024). From an information systems (IS) perspective, LLM-generated synthetic consumer data can be understood as a form of AI-enabled information generation, because generative AI systems produce new, meaningful content by drawing on patterns learned from large corpora that may include consumer reviews, social media posts, blogs, and other forms of consumer-generated text (Feuerriegel et al., 2024; Brand et al., 2023).

The focus of this research is first on showing how LLMs can be used to generate synthetic consumer data for creative and interpretive marketing research tasks, We use projective techniques as the methodological testbed because they elicit open-ended consumer associations rather than direct ratings or choices, making them useful for evaluating whether LLMs can generate richer forms of consumer insight. Tourism destinations provide an appropriate empirical context because they evoke factual knowledge, emotions, cultural imagery, and symbolic associations, all of which are central to projective consumer insight.

Accordingly, the major research questions studied are as follows: First, can LLMs generate synthetic consumer responses to projective techniques that resemble responses provided by human participants and give equivalent insights? Second, what metrics are appropriate for evaluating the similarity, diversity, and interpretive richness of LLM-generated projective data? Third, how do prompting strategies, seed examples, and model parameters affect the quality of generated responses? Fourth, do different LLMs, and different generations of LLMs, vary in their ability to generate projective consumer data?

The remainder of the paper is structured as follows. First, we provide an overview of the use of LLMs for generating synthetic data for consumer and market research. We then replicate a set of consumer data challenges using a range of LLM prompts, from basic concise prompts to more complex prompts that are "seeded" with real consumer data. We utilize a range of text-analytic measures to look at the characteristics of LLM output and compare this with the source human output. We utilize these comparisons and associated empirical work to draft recommendations for practitioners on using LLMs to create projective or other creative data.

## BACKGROUND

Since ChatGPT was launched in late 2022 employing LLMs that provide capabilities far beyond those of traditional predictive or analytic AI systems, including language generation, summarization, ideation, and simulated dialogue, generative AI has significantly impacted a wide range of fields. In marketing, it has been used to create social media content, advertisements, press releases, and branding materials; to create AI-powered customer service assistants; and to help optimize consumer experiences (e.g., Kshetri et al., 2024). A key use of generative AI is to improve the marketing research process, by making sense of large amounts of unstructured data, generating ideas for new products or services, and even generating synthetic marketing research data (Grewal et al., 2025; Sarstedt et al., 2024).

This generation of synthetic data is the focus of this work. It is likely that consumer data such as online reviews, social media posts, and blog posts are part of the training data for large consumer-scale LLMs, such as ChatGPT, and thus contain information on consumers' stated preferences (Brand et al., 2023). Given the high costs of gathering primary data on preferences and beliefs of real consumers (Market Research Society, 2024), the ability to efficiently elicit consumer beliefs and preferences and recover valid distributions of consumer meanings, preferences, and heterogeneity via LLM prompting and LLM-created synthetic datasets could offer substantial practical value to market researchers.

Work on utilizing LLMs to generate consumer data replicates the diversity of marketing research techniques and includes work on generating synthetic datasets to make choices in conjoint studies (Brand et al., 2023), replicating academic survey research in services (Bickley et al., 2025), examining if LLMs have similar risk-reward trade-offs to humans (Goli & Singh, 2024), and answering a range of structured and unstructured survey questions (Arora et al., 2025).

This research extends prior work on LLM-generated consumer data by examining whether LLMs can generate responses to projective marketing research techniques. Unlike structured survey items or choice tasks, projective techniques ask respondents to interpret, complete, or react to relatively ambiguous stimuli, thereby revealing associations, meanings, emotions, and latent perceptions that may not be captured through direct questioning (Boddy, 2005; Bond & Ramsey, 2010; Donoghue, 2000). Projective techniques, innovated by psychologists to gain deep insight into the human condition (Lilienfeld et al., 2000), have long been used by researchers looking to gain insight into consumer behavior. Early marketing applications demonstrated how indirect tasks could reveal associations that conventional surveys might miss. For example, Haire's (1950) classic shopping-list study showed how consumers' descriptions of a hypothetical shopper could reveal latent perceptions of instant coffee users, while Zober (1956) described sentence completion tasks and creative writing exercises for eliciting responses to retail and consumption scenarios. In a consumer behavior setting, the type of stimuli and level of abstraction of the stimuli can be quite varied and projective techniques run the gamut from simple ordering techniques to complex multi-step creative endeavors (Bond & Ramsey, 2010; Pich et al, 2015; Zayer et al. 2024).

This makes projective techniques a demanding testbed for synthetic data generation. If LLM-generated responses are to be useful in this context, they must do more than approximate aggregate preferences; they must reproduce patterns of association, diversity, abstraction, and interpretive richness found in human-generated data. Given the substantially different processes underlying human cognition and LLM text generation (Park et al., 2023; Sarstedt et al., 2024), the results should help us further understand some of the differences between human and LLM cognition for tasks that include both knowledge extraction and creative components.

Although the consumer information embedded in LLM training corpora is unlikely to be representative of any clearly defined population (Sarstedt et al., 2024), by altering parameters that control the diversity of the response distribution and the selection of responses, such as the temperature and top-p parameters (Renze et al., 2024), and by judiciously employing different prompting strategies (Yu et al., 2023) and personas (Ge et al., 2024), one can tune the LLM data generation process by benchmarking with human-generated data (Brand et al., 2023; Sarstedt et al., 2024). This benchmarking process may differ depending on the objectives of the research. If the objective is to obtain a set of quantitative measurements, for example, a product preference or a purchase likelihood, benchmarks could be a simple difference calibration, while for more creative exercises, evaluation could be multimodal and could include evaluations of the diversity and quality of generated insights.

From an information systems perspective, LLM-generated synthetic consumer data can be viewed as an emerging form of decision-support capability. Classic decision support systems (DSS) research emphasizes support for semi-structured managerial problems (e.g. Little, 1970) and the interaction among data, models, and dialogue or interface components (Sprague, 1980), a perspective that remains central to DSS research in information systems (Arnott & Pervan, 2014). This framework can be mapped onto LLM-generated synthetic data: the training corpus provides the underlying informational base, the LLM functions as the generative model, and prompting strategies serve as the dialogue or interface mechanism through which researchers guide system output. In the present setting, the LLM, sampling parameters, prompts, and human examples function as configurable elements of an AI-enabled data-generation process, consistent with the broader evolution of DSS technologies toward more flexible, analytics-based, and intelligent systems (Shim et al., 2002), consistent with recent IS work on agentic IS artifacts and generative AI as an active participant in human-AI value creation and collaborative innovation (Baird & Maruping, 2021; Berente et al., 2021; Fügener et al., 2022; Wessel et al., 2025).

To help clarify the current knowledge base on synthetic data for consumer research, we summarize selected work on synthetic data generation for consumer research in Table 1.

**Table 1: Summary of Literature on Synthetic Data Generation in Consumer Research**

| Paper | Data domain | LLM task | Contribution to synthetic-data literature |
|---|---|---|---|
| Arora et al. (2025) | Marketing research process | Human–LLM hybrid research design | Develops a broader framework for integrating LLMs into marketing research workflows, including ideation, survey design, and synthetic respondents. |
| Bickley et al. (2025) | Services and customer experience | Comparison of human and synthetic service research data | Provides evidence on the potential for synthetic data to reproduce patterns in service research contexts. |
| Bisbee et al. (2024) | Public opinion and political survey data | Synthetic survey respondents | Shows that synthetic response can approximate human response patterns but highlights the risks of treating LLM survey responses as a direct replacement for human responses. |
| Brand et al. (2023) | Consumer preferences and product choice | Synthetic respondents for conjoint-style choice tasks | Demonstrates the ability of LLMs to generate structured choice data and approximate consumer preference patterns. |
| Estevez et al. (2025) | Beer consumption and purchase funnel | LLM replication of consumer survey responses | Demonstrates the use of multiple LLMs to reproduce structured market research survey data in a consumer packaged goods context. |
| Goli & Singh (2024) | Behavioral decision-making | LLM responses to risk/reward and intertemporal choice tasks | Shows how LLM-generated responses can be benchmarked against established human decision-making behavior. |
| Grewal et al. (2025) | Marketing strategy and customer experience | Conceptual/review article on GenAI in marketing | Positions generative AI as a transformative technology for marketing activities, including content creation, customer interaction, innovation, and research. |
| Imschloss et al. (2025) | Sensory service research | LLM-generated sensory service responses and evaluations | Extends the use of LLM-generated data to sensory and experiential service contexts, where responses involve subjective impressions and qualitative interpretation. |
| Sarstedt et al. (2024) | Marketing research methodology | Synthetic data and LLM-based research | Provides guidelines for using LLM-generated “silicon samples” in consumer and marketing research, while identifying key methodological opportunities and validity concerns. |
| Sarstedt et al. (2026) | Marketing research practice | Practitioner guidelines for silicon samples | Translates the silicon-samples literature into practical guidance for using LLM-generated respondents in marketing research, including prompting, sampling, and validation considerations. |
| Viglia et al. (2024) | Tourism research | Synthetic data generation for tourism research | Discusses the potential for LLM-generated synthetic data in tourism, particularly for early-stage tasks such as pretesting snd stimuli generation and evaluation, while cautioning against using synthetic data in main studies. |

Prior research therefore provides both encouragement and caution. Studies such as Brand et al. (2023), Goli and Singh (2024), Arora et al. (2025), and Estevez et al. (2025) suggested that LLMs can generate useful synthetic responses for structured choice tasks, behavioral decision tasks, marketing research workflows, and some consumer survey settings. At the same time, Bisbee et al. (2024) showed that synthetic respondents may perform poorly as substitutes for human survey respondents, underscoring the need for task-specific validation. More domain-specific work has begun to examine synthetic data in tourism and experiential contexts. Viglia et al. (2024) evaluated the use of LLMs for stimuli generation, and stimuli evaluation in tourism research, while Imschloss et al. (2025) examined whether LLMs can generate contextually appropriate responses to sensory service cues, such as color, visual shape, scent, and music, capture how these cues combine to shape consumer perceptions. Building on this work, we examine whether LLM-generated data can support projective techniques in tourism destination research. This setting provides a demanding test because projective techniques require more than the reproduction of structured preferences or survey responses; they require the generation of associations, emotions, symbolic meanings, and creative interpretations in response to ambiguous stimuli.

## METHODOLOGY

Blinded Authors (Authors withheld for blinded review, 2025) contains a substantive primary data collection using projective techniques for a city tourism scenario for five U.S. cities. Respondents were asked to give gender, previous visit information, and complete association and completion tasks to elicit insights into consumers' perceptions of these cities. The resulting data were analyzed using both qualitative (thematic analysis) and quantitative (statistical testing, correspondence analysis) methodologies.

In the current study, we replicate this data-collection process using LLM-generated synthetic consumer responses. We examine how LLM model choice, model settings, and prompt design affect the characteristics and quality of the generated data, allowing us to explore the research questions developed in the Introduction section.

The human benchmark data were drawn from the primary data collection reported in Blinded Authors (2025). Participants were recruited from a student sample pool at a large U.S. state university in the Southeast of the United States of America. This sample was appropriate for the present tourism destination setting because the study focused on perceptions of major U.S. cities among college students, a consumer group for whom leisure travel and post-graduation relocation are salient.

Participants completed a paper booklet containing projective tasks for five U.S. tourist destinations: Las Vegas, Los Angeles, Nashville, New Orleans, and New York. The booklet was administered in four randomized variants that varied the order of the cities and the order of the word-association and sentence-completion tasks. Participants also reported whether they had previously visited each destination and indicated their gender; no other demographic information was collected.

The human sample consisted of 173 respondents across four sessions. Among the 172 respondents who completed the city-visit questions correctly, 27 had visited one of the cities, 65 had visited two, 49 had visited three, 20 had visited four, and 6 had visited all five. Only 5 respondents had not visited any of the cities.

The projective tasks used in the present comparison included word association and sentence completion tasks, both commonly used in consumer research and which are often used in tandem to give complementary insights (e.g., Alencar et al., 2021; Chen et al., 2022; Eldesouky et al., 2015; Grougiou & Pettigrew, 2011). For the word-association task, respondents were asked to provide up to four “impressions, words, images, associations, or feelings” that came to mind for each city. The sentence-completion task presented a visual dialogue between two people with sentence completion bubbles. In the positive condition, the first speaker stated, “I heard that you had a great time in [city],” and respondents completed the reply beginning “Yes, I…” In the negative condition, the first speaker stated, “I heard that you had a terrible time in [city],” followed by the same response stem. The order of the positive and negative prompts and the gender placement of the speakers in the image were randomized.

For the LLM comparison, we tested multiple different LLMs, multiple temperature settings, and multiple different prompting strategies. These are summarized in Table 2.

**Table 2: Experimental Variables**

| Variable | Description | Levels |
|---|---|---|
| City | The tourism city to be tested. | Las Vegas, Los Angeles, Nashville, New Orleans, New York City |
| ProjectiveTask | The projective technique method to be used. | WA (Word Association)<br>SC- (Sentence Completion Negative)<br>SC+ (Sentence Completion Positive) |
| Model | The LLM model used including the version. | ChatGPT 3.5<br>ChatGPT 4.1<br>ChatGPT 5.1<br>Gemini 2.5 Flash<br>Grok 4.1<br>Mistral Medium 3.1 |
| Temperature | The temperature setting for the model. | 1.0, 1.2, 1.4, 1.6, 1.8 |
| PromptingStrategy | The prompting strategy for retrieving the ChatGPT data. | Basic<br>Extended<br>ExtendedProb<br>ExtendedProb10Shot<br>ExtendedProb50Shot |

The five cities replicated the five cities included in the original tourism study. Each of the cities had results for Word Associations (WA), Sentence Completions Negative (SC-), and Sentence Completions Positive (SC+) using the tasks described above. There were $r$ = 173 responses (with no more than 3 missing responses for each city). To aid comparisons, we kept a similar number of r = 173 responses for each combination of experimental parameters. The focus when choosing LLMs was to include all three generations of the most prominent LLM (ChatGPT), but also to include a selection of non-OpenAI LLMs, to give a sampling of the breadth of available LLMs. Cost considerations were important, as given the number of runs needed for the experiment, some expensive "high-level" models were not feasible, so model choice was a trade-off between performance and cost. For example, for Gemini, we picked the mid-level "Flash" model, which was situated between the basic "Flash-Light" model and the expensive "Premium" model.

The temperature parameter controls the stochasticity of LLM output by rescaling the probability distribution over candidate tokens before sampling (Davis et al., 2024; Renze et al., 2024). Lower temperature values concentrate probability mass on the highest-probability tokens, producing more stable and deterministic responses, whereas higher values flatten the distribution and increase the likelihood of less probable tokens, producing more varied and creative responses. A related sampling parameter is top-p, or nucleus sampling, which restricts sampling to the smallest set of candidate tokens whose cumulative probability reaches a specified threshold (Holtzman et al., 2019; Min et al., 2022; Renze et al., 2024).

In the present study, we varied temperature while holding top-p to default settings because OpenAI's API documentation (Microsoft, 2026) recommends adjusting either temperature or top-p, but not both, as changing both simultaneously can make the source of variation in generated responses more difficult to interpret. In addition, temperature provides a continuous way to alter response diversity, while top-p changes the candidate sampling set more directly.

We give an illustration of how the temperature setting works in Figure 1. The top panel shows the response distribution with the percentile of responses, going from most likely to least likely on the x axis. The y-axis shows response probability on a logarithmic scale, allowing both common high-probability responses and rare low-probability responses to be visible in the same plot. The lower panels of the figure show how the distribution is adapted for different temperature settings, with the dashed lines giving the base probabilities and the solid lines giving the probabilities modified for temperature. Lower temperatures give a steeper distribution curve which results in more deterministic sampling. Higher temperatures give flatter probabilities, which results in a higher diversity of responses.

**Figure 1: Illustrative LLM Response Distributions Adapted for Temperature**

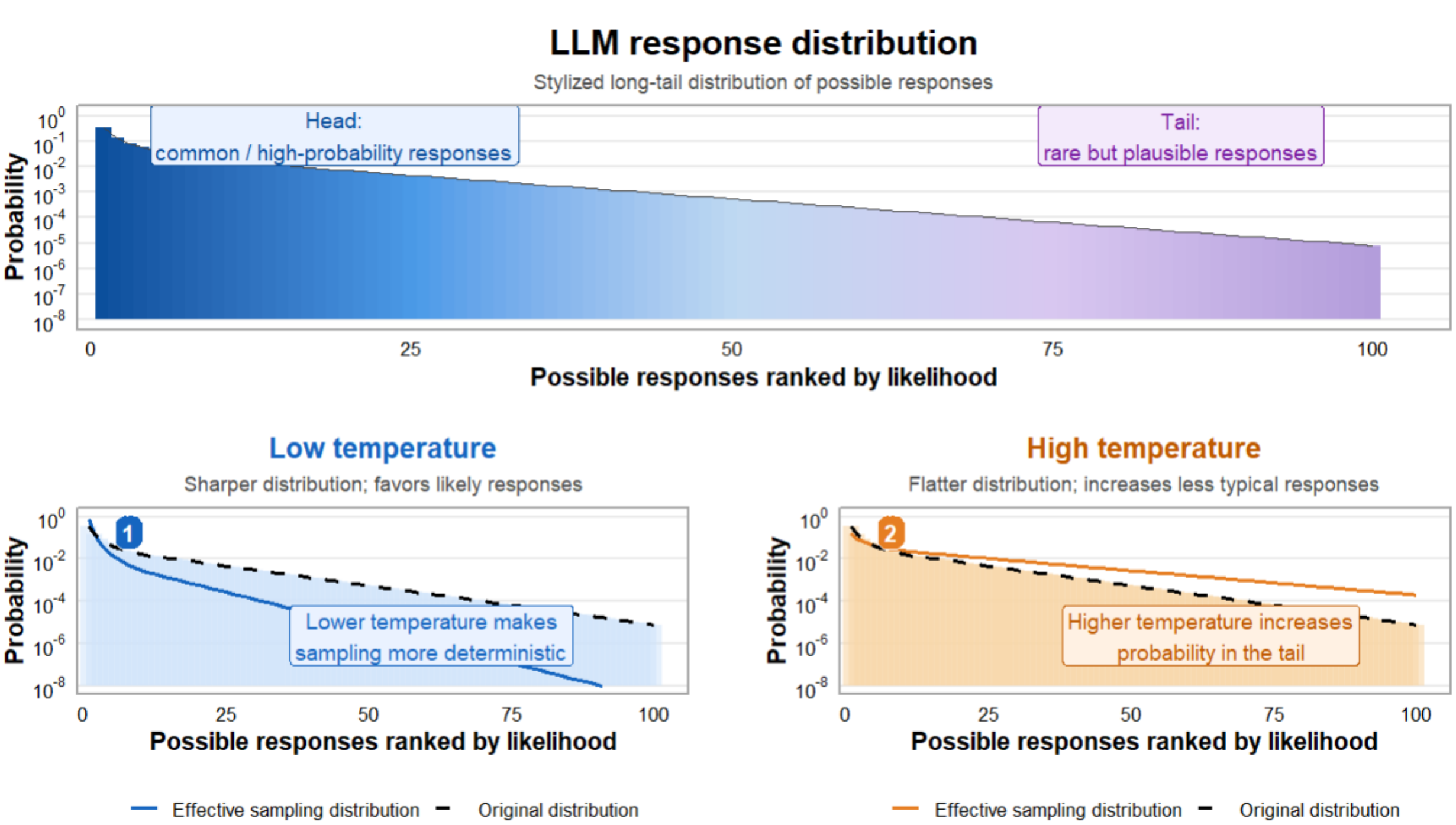

The temperature setting for the OpenAI LLM can range from 0 to 2. Initial pretests determined that for the OpenAI platform temperatures below 1.0 were almost completely deterministic. Conversely, pretests showed that very high temperature settings could produce degenerate outputs rather than useful response diversity. At temperatures of approximately 1.6 and above, some responses began coherently but then drifted into unusual symbols, multilingual fragments, or semantically incoherent text, especially for older models. This is consistent with decoding-induced text degeneration, in which high-temperature sampling increases the likelihood of drawing from the unreliable low-probability tail of the token distribution (Holtzman et al., 2019). For the OpenAI models, the proportion of degenerate responses was typically quite low (<5%), so we just requeried the LLM for clearly degenerate text. However, we capped the temperature at 1.8, as values above this value gave an unacceptably high proportion of degenerate responses. The Grok advanced reasoning models seemed to have a particular problem with this degeneracy, which made data collection infeasible, so for Grok we picked the basic "nonreasoning" model. All non-OpenAI models were run using the same temperature values as the OpenAI models, using the values specified in Table 2 [1].

**LLM Prompting**

The overall rationale behind the prompting strategies was to begin with the prompt used in the human data collection, but then implement a range of alternative prompt designs intended to improve the diversity and substantive content of the LLM-generated responses. This was necessary because prompts written for human respondents may not produce equivalent responses when given to LLMs. Unlike human respondents, LLMs generate text by conditioning on the prompt and sampling from learned language distributions; therefore, changes in task framing, contextual information and examples can affect the resulting responses. This design is consistent with prompt-engineering research showing that LLM outputs are sensitive to prompt wording and structure (White et al., 2023; Sahoo et al., 2024; Schulhoff et al., 2024), and with work showing that model performance can depend substantially on the quality of the prompt used (Zhou et al., 2022).

[1] The one exception was the Mistal model, where temperature was multiplied by 0.75, due to a lower effective upper temperature bound.

The OpenAI API allows the splitting of the prompt into system and user prompts. The system prompt allows the LLM to store contextual information separately from the prompt question. Initial tests did not find much difference when splitting the prompt, and as not all APIs implemented this feature in the same way we put all the text in a single prompt. Gender information was collected for the human data collection, and gender was found to significantly impact responses, so we randomly sampled a *gender* variable, using a probability distribution derived from the frequency of genders in the survey (0.46 male, 0.54 female).

The first prompt, the Basic prompt was adapted from the human data collection. The prompt for the word association is as follows.

*Imagine that you are a {gender} college student from the South East of the U.S. You are planning a long weekend city break away with friends in {city}. Please give four different descriptive adjectives or short phrases that you think of in connection with {city}. Put one response on each line and do not add any additional explanation.*

The last sentence was needed to ensure consistent response structure and to ensure that the LLM did not add any clarifying information to answers.

The prompt for the sentence completion tasks is given below. For negative completion, *terrible* and *negative* are used as descriptive adjectives and for positive completion, *great* and *positive* are used as descriptive adjectives. These match the adjectives used for the human data collection.

*Imagine that you are a {gender} college student from the South East of the U.S. You are planning a long weekend city break away with friends in {city}. You are conversing with a friend. Your friend says "I heard that you had a {terrible/great} time in {city}.You respond with "Yes, I........". Please complete this sentence, thinking about a potential scenario where you*

*could have a {negative/positive} terrible time in {city}. Put the response on a single line and do not add any additional explanation.*

The Extended prompt was designed to elicit more creativity and to generate more descriptive responses. For word association, the following was added.

*Be creative and think about positive and negative aspects of {city}, along with {city}'s attractions, sights, and events.*

For sentence completion, the following was added.

*Be creative and think about different aspects of {city}, along with {city}'s attractions, sights, and events.*

The prompts were similar, but for sentence completion the base prompt already specified polarity, while for word association adding an instruction to think about both "positive and negative" aspects of the city was used to help generate more diverse insights.

The ExtendedProb prompt includes the following additional instruction for both word association and sentence completion tasks.

*Generate your responses, sampled from the full distribution of possible responses.*

This prompt was designed to increase the diversity of responses by implementing verbalized sampling (Franzen, 2025; Zhang et al., 2025), a prompting strategy that encourages the model to consider a broader range of plausible outputs rather than defaulting to the most typical response.

The final two prompts included randomly selected answers from the data collection. This approach is often called few-shot prompting (Brown et al., 2020; Sahoo et al., 2024) and the rationale here is that combining human and LLM data collection can magnify the advantages of both, with cost savings from requiring fewer human respondents, while using human examples to guide the LLM toward the style, content, and level of specificity observed in the target population and still allowing the model to generate new synthetic responses. We added prompts that extended the prompt further with 10 shot (ExtendedProb10Shot) and 50 shot (ExtendedProb50Shot) prompting. The prompt is below and it was then followed by the human output for each of the randomly selected examples.

*The following examples are provided to give an example of content and format. However, please use your own corpus to generate your response.*

## MAIN RESULTS

The results address our initial research questions: whether LLMs can generate projective consumer responses that approximate human responses, how such responses should be evaluated, how prompt design and model parameters shape the generated data, and whether different LLMs vary in their ability to produce useful projective data. We calculated a range of metrics for the $r = 173$ responses for each combination of city and experimental setting (i.e., ProjectiveTask, Model, Temperature, and PromptingStrategy) and also for the human responses.

### Defining Textual Response Metrics

We calculated both basic commonly used descriptive metrics for text characteristics and also metrics that measure the diversity of responses. These metrics are described below.

Assume that each response is tokenized by splitting the text into words and removing punctuation. Let $n_i$ be the number of word tokens in response $i$ and let $W$ be the set of all potential words. Table 3 defines the basic textual characteristics used to summarize response length, vocabulary size, word length, and stopword usage. These measures are standard descriptive features in corpus and lexical-richness analysis (e.g., Tweedie & Baayen, 1998; Baayen, 2001).

**Table 3: Basic Textual Characteristics**

| Measure | Description | Formula |
|---|---|---|
| AvgWordCount | The sum of the number of word tokens in each individual response divided by the number of responses. This captures the verbosity of the responses | The total number of word tokens can be calculated by averaging the number of word tokens across responses. $N = \sum_{i=1}^{r} v$ $\text{AWC} = \frac{1}{r} N$ |
| VocabSize | This is the number of unique words used for a city. It gives a simple indication of the overall diversity of the responses. | The set of words that occur once is defined as V. $V = \{w \in W \mid n_w > 0\}$ , where *w* is any potential response word. The VocabSize is the size of this set. $VS = \lvert V \rvert$. |
| AvgWordLen | The average length of the words. If length (*w*) is the length function for a word, then average word length is the length of each word multiplied by the probability of that word occurring. | First, define $n_{i,w}$ as the number of instances of word w in response *i*. The probability of the word can be calculated as follows. $p_w = \frac{\sum_{i=1}^{r} n_{i,w}}{\sum_{i=1}^{r} N_i}$ $AWL = \sum_{w \in V} p_w \times \text{length(w)}$ |
| pStopWords | The proportion of words that are stop words (from a standard list of stop words). This measures the proportion of all tokens that are stopwords. A high stopword rate may indicate more sentence-like or verbose responses, while a lower stopword rate may indicate more list-like responses with content words. | Let *S* be the set of all stop words from a standard stop word corpus. The probability of a word being a stop word is the sum of probabilities of all words in this corpus. $p_{SW} = \sum_{w \in S} p_w$ |

We derived more complex measures of vocabulary breadth, word diversity, and concentration/repetition from these basic metrics. These metrics are given in Table 4.

Vocabulary-breadth measures such as type-token ratio and hapax usage capture the range of terms used in the corpus. Entropy-based measures evaluate the evenness of the word-frequency distribution, a common approach to quantifying diversity in information theory and text analysis (Shannon, 1948). Concentration measures such as Simpson concentration (Simpson, 1949) and Yule's K (Yule, 1944) complement entropy by identifying whether responses are dominated by a small number of repeated terms.

**Table 4: Measures of Breadth, Diversity, and Concentration**

| Measure | Description | Formula |
|---|---|---|
| p(Top 10) (Baayen, 2001) | The proportion of words that are part of the top 10 most common words. Higher values indicate that responses are concentrated around a small number of recurring words, whereas lower values indicate a more dispersed vocabulary. | Let $T_{10}$ be the set of the ten most frequent words across all responses. $p_{T10} = \sum_{w \in T_{10}} p_w$ |
| LexDiver (Bestgen, 2024; Tweedie & Baayen, 1998) | The ratio of unique words to total words. Higher values indicate that a larger proportion of the words are distinct. | $LD = \frac{VS}{N}$ |
| Entropy (Shannon, 1948) | Shannon entropy measures how evenly word usage is spread across the vocabulary. Higher entropy means the word distribution is more even and diverse; lower entropy means a few words dominate. | $H = -\sum_{w \in V} p_w \, log_2(p_w)$ Entropy can be normalized on scale of [0,1] using the log of the vocabulary size as a scaling constant. $NH = \frac{H}{log_2(VS)}$ |
| Hapax (Baayen, 2001) | A hapax word is a word that appears only once. A higher hapax rate suggests a broader or more varied vocabulary. | $HP = \frac{\sum_{w \in V} \mathbf{1}\{n_w = 1\}}{N}$ |
| Yule's k (Yule, 1944) | Yule's K is a concentration metric. Higher values indicate that a small number of words are used repeatedly. Lower values indicate greater lexical variety. | $K = 10000 \times \frac{\sum_{w \in V}(n_w^2 - n_w)}{N^2}$ |
| Simpson (Simpson, 1949) | The Simpson measure is also a concentration metric. Higher values mean the text is more dominated by a small number of frequent words. Lower values mean word usage is more dispersed. | $SC = \sum_{w \in V} p_w^2$ |

In summary, the basic volume and style measures capture differences in verbosity and response format, while measures for linguistic diversity and concentration capture the breadth and repetition of the response text.. Because lexical-diversity metrics measure different aspects of lexical diversity can be sensitive to text length, we report multiple complementary measures rather than relying on a single index (Baayen, 2001; Bestgen, 2023).

### Basic Linguistic Characteristics

We took the results from all runs of the experiment. For each response, we split the response into words and then stemmed (e.g., Singh & Gupta, 2016) the words, i.e., combining similar words with different grammatical endings (e.g., fly, flies, flying, etc.). For word association tasks, the four responses were counted as separate responses. We then calculated the metrics described in the previous section for each set of responses, giving a single aggregate measurement for each combination of experimental settings and city. For each metric, we calculated regression models, with the metric as the dependent variable and the experimental conditions (city, projective task, LLM, Temperature, and prompt mode) as independent variables. The results for the basic linguistic characteristics are given in Table 5. The base levels for the categorical dummy variables are city (Las Vegas), projective task (SC-), LLM (Gemini 2.5), and prompt mode (Basic).

Making some overall conclusions, first the effect sizes for the regression show substantial explanatory power ($R^2 > 0.26$), using guidance from Cohen (1988), and range from 0.538 (VocabSize) to 0.896 (StopWord). The temperature has a significant positive relationship with the number of words ($p < .01$), the vocabulary size ($p < .001$), and a significant negative relationship with the proportion of stop words ($p < .05$). Together, this indicates that higher temperatures give longer, more diverse responses, with a smaller proportion of "standard" stop words. The extended prompt has significant positive relationships with the number of words, the vocabulary size, and the word length (all $p < .001$). The extended prompt with the additional instruction to select from the whole probability distribution has almost identical significance results, but with word length $p < .05$ rather than $p < .001$. This indicates some success in improving the diversity of responses over the basic prompt. The 10-shot and 50-shot prompting conditions remain strongly associated with larger vocabularies ($p < .001$), but their effects on word count are smaller ($p < .01$ for 50 shot and $p < .05$ for 10 shot) and their effects on word length are not significant. This suggests that including human examples may guide the LLM toward the human responses, producing a more targeted form of diversity rather than simply maximizing response breadth.

**Table 5: Regression Models for Basic Linguistic Characteristics**

| Term | NoWords | VocabSize | WordLen | StopWord |
| --- | --- | --- | --- | --- |
| Intercept | 21.047$^{***}$<br>(1.702) | -58.850$^{\bullet}$<br>(32.847) | 3.915$^{***}$<br>(0.039) | 0.506$^{***}$<br>(0.008) |
| CityLosAngeles | 1.052<br>(0.846) | -17.144<br>(16.322) | 0.136$^{***}$<br>(0.019) | 0.020$^{**}$<br>(0.004) |
| CityNashville | -1.276<br>(0.846) | -32.033$^{*}$<br>(16.322) | 0.211$^{***}$<br>(0.019) | -0.036$^{***}$<br>(0.004) |
| CityNewOrleans | 1.336<br>(0.846) | -45.133$^{**}$<br>(16.322) | 0.060$^{**}$<br>(0.019) | -0.020$^{***}$<br>(0.004) |
| CityNewYorkCity | -0.889<br>(0.846) | 15.680<br>(16.322) | -0.033<br>(0.019) | -0.032$^{***}$<br>(0.004) |
| ProjTaskSC+ | -0.848<br>(0.655) | -208.280$^{***}$<br>(12.643) | 0.128$^{***}$<br>(0.015) | -0.051$^{***}$<br>(0.003) |
| ProjTaskWAS | -35.744$^{***}$<br>(0.655) | -292.939$^{***}$<br>(12.643) | 1.800$^{***}$<br>(0.015) | -0.374$^{***}$<br>(0.003) |
| LLMgpt-35 | -7.490$^{***}$<br>(0.926) | -203.843$^{***}$<br>(17.880) | -0.196$^{***}$<br>(0.021) | 0.015$^{***}$<br>(0.004) |
| LLMgpt-41 | 2.606$^{**}$<br>(0.926) | -11.557<br>(17.880) | -0.182$^{***}$<br>(0.021) | 0.014$^{**}$<br>(0.004) |
| LLMgpt-51 | 26.562$^{***}$<br>(0.926) | 198.722$^{***}$<br>(17.880) | -0.278$^{***}$<br>(0.021) | 0.050$^{***}$<br>(0.004) |
| LLMgro-41 | 4.345$^{***}$<br>(0.926) | -74.165$^{***}$<br>(17.880) | -0.104$^{***}$<br>(0.021) | -0.013$^{***}$<br>(0.004) |
| LLMmis-ml | 19.648$^{***}$<br>(0.926) | 282.933$^{***}$<br>(17.880) | -0.460$^{***}$<br>(0.021) | 0.084$^{***}$<br>(0.004) |
| Temperature | 2.769$^{**}$<br>(0.945) | 352.461$^{***}$<br>(18.248) | 0.030<br>(0.021) | -0.010$^{*}$<br>(0.004) |
| PromptEx | 18.195$^{***}$<br>(0.846) | 366.618$^{***}$<br>(16.322) | 0.101$^{***}$<br>(0.019) | 0.003<br>(0.004) |
| PromptExProb | 12.867$^{***}$<br>(0.846) | 320.651$^{***}$<br>(16.322) | 0.045$^{*}$<br>(0.019) | 0.017$^{***}$<br>(0.004) |
| PromptExProb10Shot | 1.634$^{\bullet}$<br>(0.846) | 165.633$^{***}$<br>(16.322) | 0.018<br>(0.019) | 0.003<br>(0.004) |
| PromptExProb50Shot | 2.501$^{**}$<br>(0.846) | 174.611$^{***}$<br>(16.322) | -0.025<br>(0.019) | 0.013$^{**}$<br>(0.004) |
| $R^2$ | 0.746 | 0.538 | 0.895 | 0.896 |
| | $^{***}p < 0.001$, $^{**}p < 0.01$, $^{*}p < 0.05$, $^{\bullet}p < 0.10$ | | | |

There are mixed results for the city fixed effects, with some significant results. This indicates that different destinations elicit somewhat different response patterns. For the projective task type, the word association task has a significantly (all $p$ <.001) lower number of words, vocabulary size, and stop word proportions, which is to be expected because respondents were asked to provide descriptive words or short phrases rather than complete sentences. However, word association also has significantly higher average word length ($p < .001$), suggesting that the shorter responses may contain more content-heavy terms. Interestingly, despite almost identical tasks, positive sentence completion has significantly lower vocabulary size ($p < .001$), lower stop word proportion ($p < .001$), and higher average word length ($p < .001$) than negative sentence completion. This suggests that positive and negative destination prompts elicit different linguistic patterns even when the task format is otherwise similar.

There were also significant differences across LLMs. Relative to the Gemini 2.5 baseline, ChatGPT 5.1 and Mistral Medium 3.1 produced substantially longer responses and larger vocabularies ($p < .001$), whereas ChatGPT 3.5 produced significantly shorter responses and a smaller vocabulary ($p < .001$). These patterns show that model choice has a strong effect on the form of the generated data. For projective consumer research, longer responses may be more useful for exploratory idea generation than for closely reproducing the brevity and structure of human responses. Interestingly, the Gemini 2.5 baseline produced significantly longer average word lengths than all other models ($p < .001$). Stop word usage also differed by model: Gemini had a significantly lower proportion of stop words than the ChatGPT models ($p < .01$) and Mistral Medium 3.1 ($p < .001$), but a significantly higher proportion of stop words than Grok 4.1. These patterns suggest that model differences extend beyond response length and vocabulary size to response style and linguistic structure.

One of the initial research questions was whether the LLM-generated responses resembled those from the human sample. However, some of the summary statistics, such as overall vocabulary size and entropy, are defined at the corpus level rather than at the individual-response level. We therefore calculated these metrics across all (n = 173) responses for each LLM condition-city combination and, separately, for each corresponding human city-task combination. Because this produces a single aggregate value for each human benchmark and each LLM condition, the resulting comparisons are primarily descriptive rather than inferential. Accordingly, we evaluate these summary values visually, plotting the human values against LLM values in a series of scatterplots. Because the experiment produced a large number of LLM condition-city combinations, we present a subset of conditions in comparison with the human benchmark. Guided by the regression results, we selected two contrasting generation settings for each of the LLMs: the Basic prompt with temperature 1.0 and the ExtendedProb prompt with temperature 1.8. These conditions approximate lower- and upper-diversity settings, respectively, allowing us to compare human responses with both relatively constrained and more diversity-oriented LLM outputs. Averaging across cities, this gives 12 LLM values to compare with each human value.

**Figure 2: Word Association: Basic Characteristics Human vs. LLM**

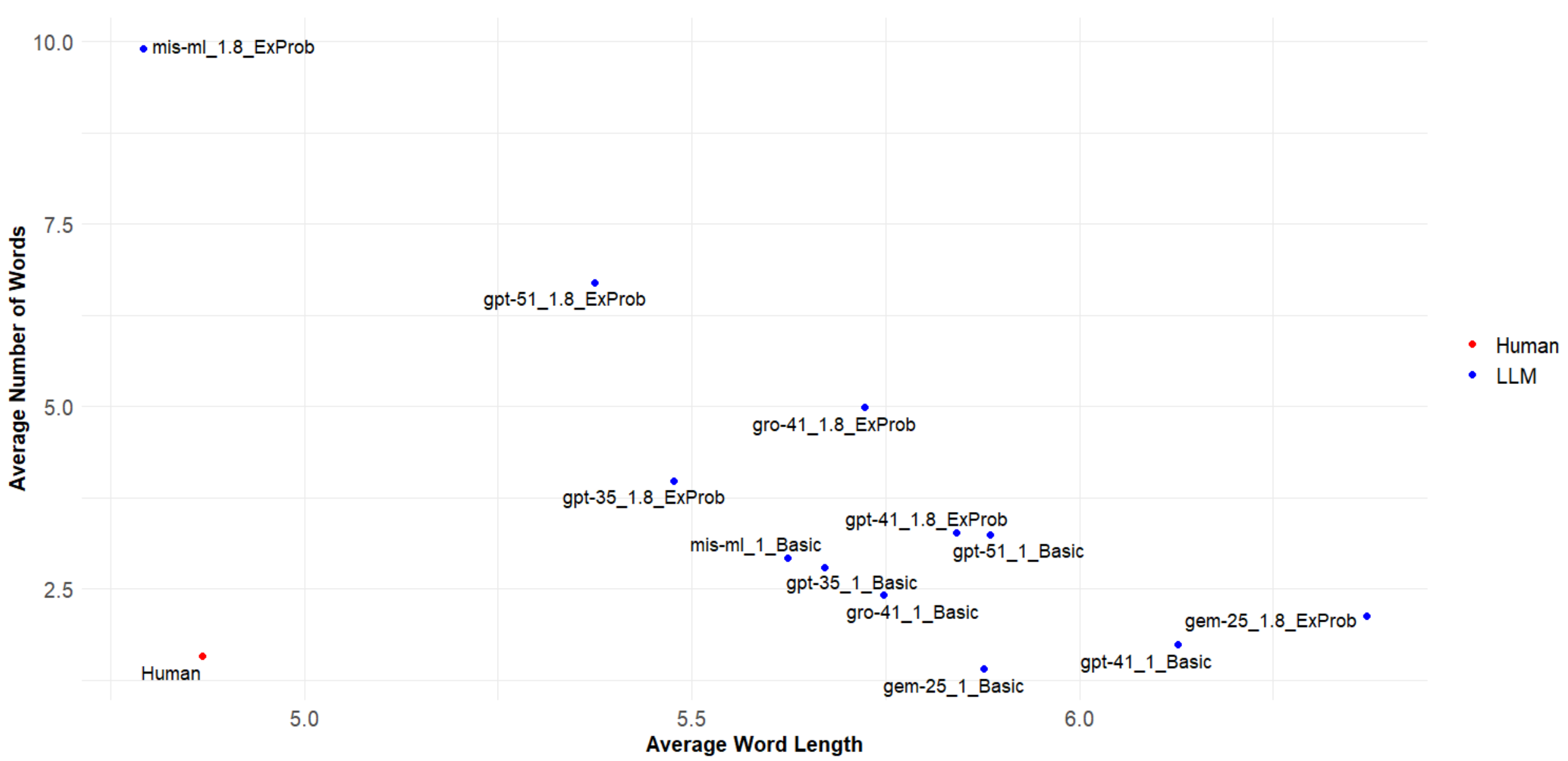

We focus on the basic characteristics of word length and number of words. Figures 2 to 4 show the average word length vs. the average number of words in each response for WA, SC+, and SC- scenarios respectively.

**Figure 3: Positive Sentence Completion: Basic Characteristics Human vs. LLM**

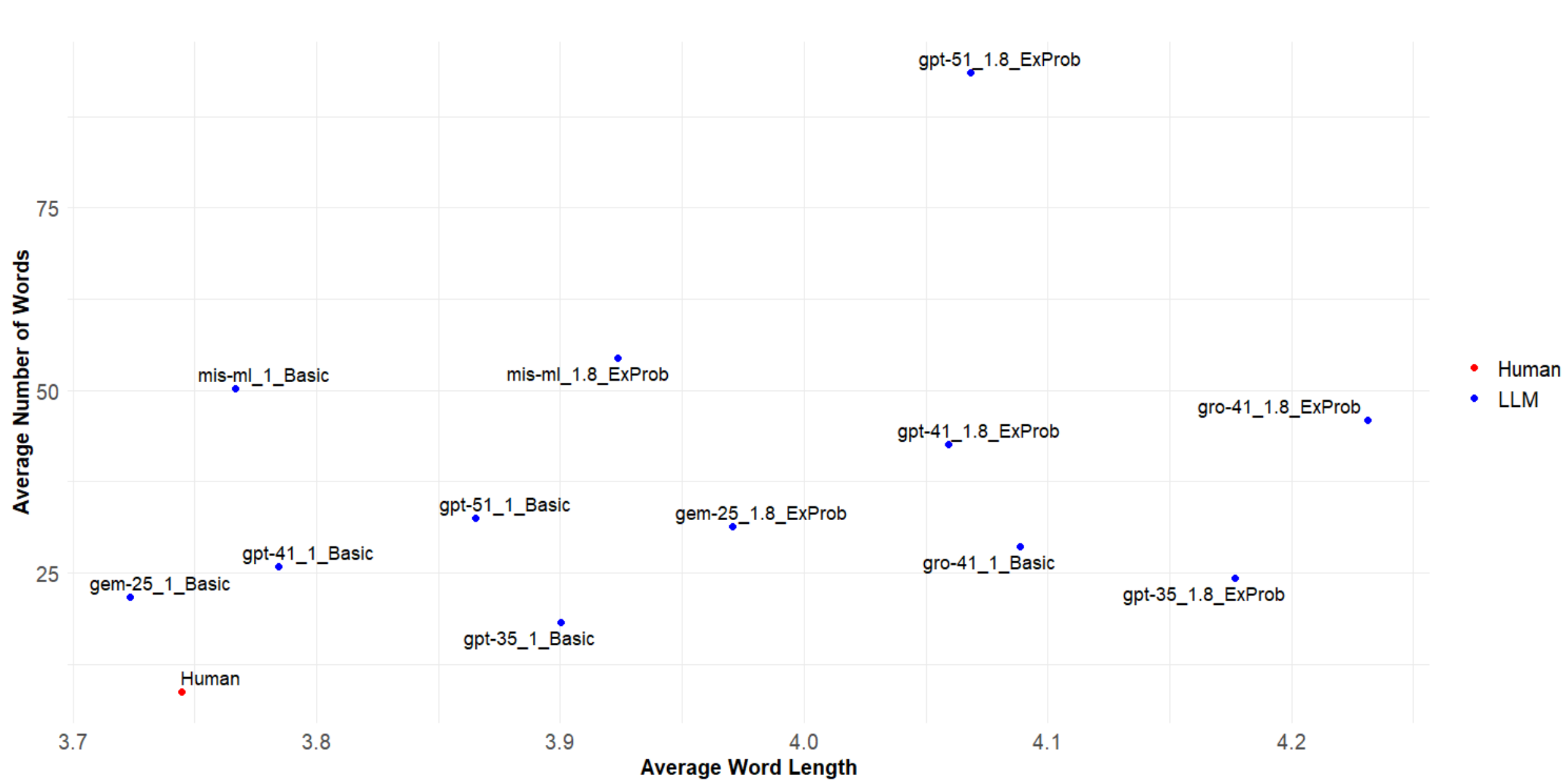

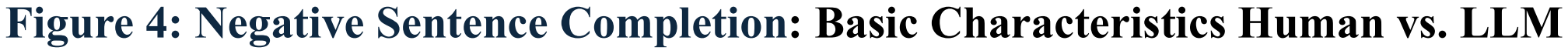

**Figure 4: Negative Sentence Completion: Basic Characteristics Human vs. LLM**

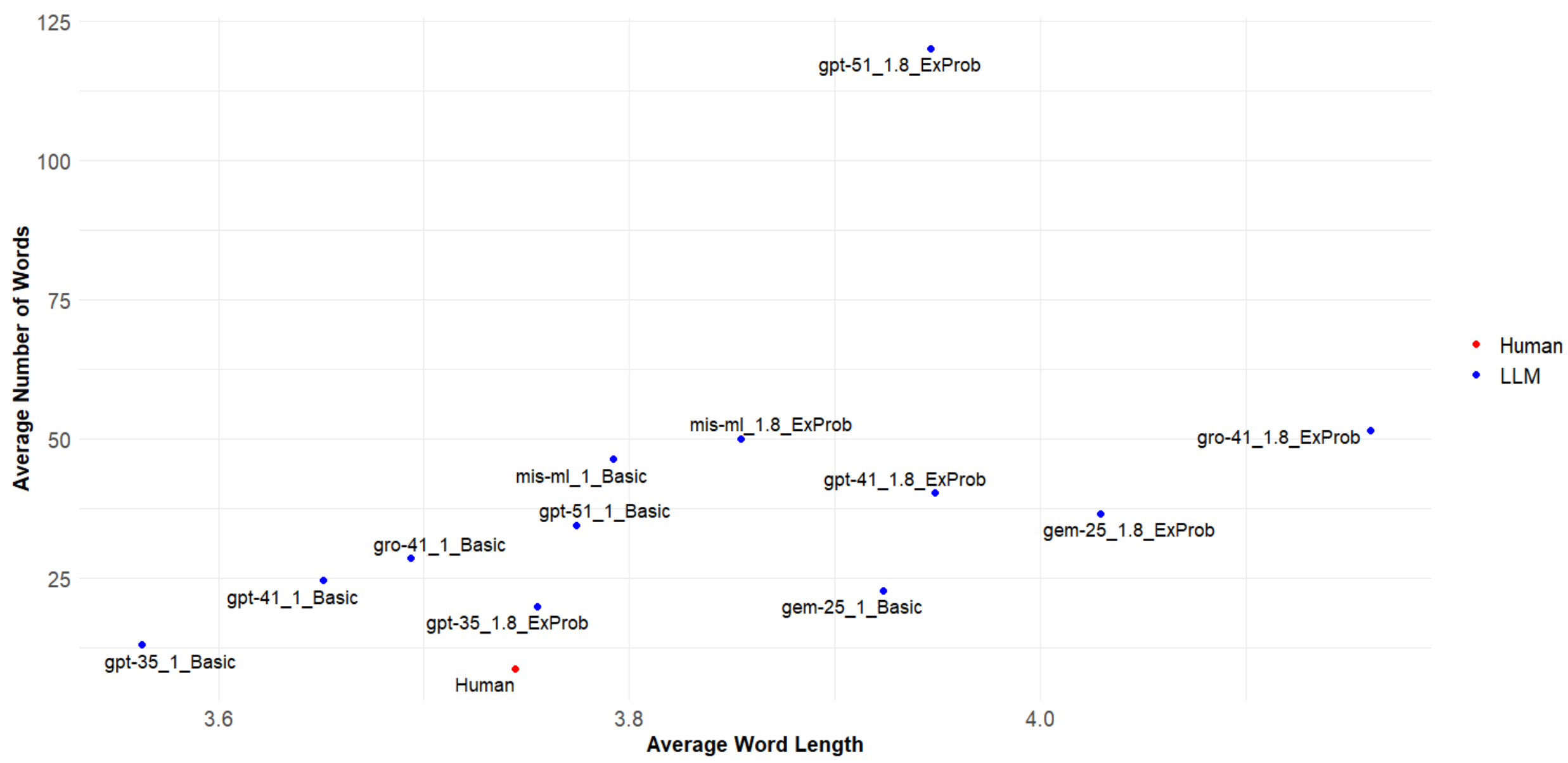

Across all three projective tasks, the human responses had the lowest, or close to the lowest, average number of words, indicating that the LLMs often generated more verbose responses than human participants. This was particularly true for extended prompts at high temperatures, which produced longer and more elaborate outputs. While these responses may be valuable for exploratory idea generation, they appear less suitable when the goal is to reproduce the brevity and succinctness of human projective responses. Average word length showed a weaker but similar pattern, with human responses below the overall average across all three tasks and second shortest for both word association and positive sentence completion.

**Table 6: Regression Models for Diversity Characteristics**

| **Term** | **Diversity** | | | **Concentration** | | |
|---|---|---|---|---|---|---|
| | **LexDiver** | **Hapax** | **Entropy** | **p(Top 10)** | **Yule's k** | **Simpson** |
| Intercept | 0.033*** (0.004) | -0.009* (0.003) | 5.371*** (0.063) | 0.506*** (0.009) | 362.796*** (9.902) | 0.037*** (0.001) |
| CityLosAngeles | -0.005* (0.002) | -0.005** (0.002) | 0.030 (0.031) | -0.008• (0.004) | -10.081* (4.920) | -0.001* (0.000) |
| CityNashville | -0.006* (0.002) | -0.003* (0.002) | -0.077* (0.031) | 0.012** (0.004) | 5.314 (4.920) | 0.001 (0.000) |
| CityNewOrleans | -0.010*** (0.002) | -0.007*** (0.002) | -0.045 (0.031) | -0.011** (0.004) | -15.127** (4.920) | -0.002** (0.000) |
| CityNewYorkCity | 0.001 (0.002) | -0.001 (0.002) | 0.182*** (0.031) | -0.028*** (0.004) | -31.235*** (4.920) | -0.003*** (0.000) |
| ExpTypeSC+ | -0.033*** (0.002) | -0.021*** (0.001) | -0.346*** (0.024) | 0.006• (0.003) | 18.153*** (3.811) | 0.002*** (0.000) |
| ExpTypeWAS | 0.033*** (0.002) | -0.019*** (0.001) | -0.582*** (0.024) | 0.064*** (0.003) | 80.436*** (3.811) | 0.008*** (0.000) |
| LLMgpt-35 | -0.042 *** (0.002) | -0.020*** (0.002) | -0.954*** (0.034) | 0.116*** (0.005) | 120.684*** (5.390) | 0.012*** (0.001) |
| LLMgpt-41 | -0.032*** (0.003) | -0.015*** (0.002) | -0.279*** (0.034) | 0.020*** (0.005) | 28.183*** (5.390) | 0.003*** (0.001) |
| LLMgpt-51 | -0.061*** (0.002) | -0.030*** (0.002) | 0.129*** (0.034) | -0.025*** (0.005) | -8.241*** (5.390) | -0.001* (0.001) |
| LLMgro-41 | -0.069*** (0.002) | -0.033*** (0.002) | -0.774*** (0.034) | 0.071*** (0.005) | 84.298*** (5.390) | 0.008*** (0.001) |
| LLMmis-ml | -0.049*** (0.002) | -0.025*** (0.002) | 0.272*** (0.034) | -0.030*** (0.005) | -18.508*** (5.390) | -0.002*** (0.001) |
| Temperature | 0.064*** (0.002) | 0.046*** (0.002) | 0.883*** (0.035) | -0.074*** (0.005) | -82.979*** (6.980) | -0.008*** (0.001) |
| ModeEx | 0.029*** (0.002) | 0.018*** (0.002) | 0.995*** (0.031) | -0.112*** (0.004) | -122.006*** (4.920) | -0.012*** (0.000) |
| ModeExProb | 0.032*** (0.002) | 0.020*** (0.002) | 1.036*** (0.031) | -0.121*** (0.004) | -131.206*** (4.920) | -0.013*** (0.000) |
| ModeExProb10Shot | 0.038*** (0.002) | 0.021*** (0.002) | 0.812*** (0.031) | -0.095*** (0.004) | -113.066*** (4.920) | -0.011*** (0.000) |
| ModeExProb50Shot | 0.038*** (0.002) | 0.019*** (0.002) | 0.865*** (0.031) | -0.101*** (0.004) | -118.363*** (4.920) | -0.012*** (0.000) |
| $R^2$ | 0.609 | 0.498 | 0.688 | 0.598 | 0.561 | 0.593 |
| | *** $p < 0.001$, ** $p < 0.01$, * $p < 0.05$, • $p < 0.10$ | | | | | |

Similar regression analyses were performed for the measures of diversity and concentration. The results are given in Table 6. The first three columns contain the results for diversity and the second three columns contain the results for concentration.

As per the basic linguistic features, the effect sizes for the regression show substantial explanatory power and range from 0.498 (Hapax) to 0.688 (Entropy). For temperature, the three diversity measures increase and the three concentration measures decrease (all $p < .001$) as the temperature increases, indicating that increasing the temperature parameter increases the diversity of prompts. All four extended prompts show significantly increases in diversity and decreases in concentration from the Basic prompt across all six metrics ($p < .001$ for all $4 \times 6 = 24$ prompt × metric combinations). Unlike with the basic linguistic features, the coefficients for the 10 and 50 shot prompting are similar to those for the non-augmented extended prompts, indicating that these prompts could have the benefit of increasing diversity while guiding the prompts toward similar style and content as the human prompts.

As with the basic linguistic characteristics, there are mixed results for the city fixed effects, with some significant results. This again indicates that different destinations have different response patterns. For projective task type, both the word-association task and the positive sentence-completion task had significantly lower values on the diversity metrics and significantly higher values on the concentration metrics than the negative sentence-completion baseline ($p < .001$ apart from SC+ and p(Top 10), which has $p < .1$ marginal significance). This suggests that task type affects not only response length, but also the breadth of associations generated. In the tourism context, the positive sentence-completion task elicits common destination benefits, such as attractions, entertainment, food, nightlife, or sightseeing, leading to more concentrated response distributions. By contrast, positive sentence-completion task may open a wider range of possible scenarios, including safety concerns, expense, crowds, weather, transportation, disappointment, or service failures, producing more varied language. Word association is also more concentrated, likely because the task encourages short, salient destination associations rather than elaborate scenarios.

Again, there were significant differences across LLMs. Relative to the Gemini 2.5 baseline, ChatGPT 3.5, ChatGPT 4.1, and Grok 4.1 showed significantly lower diversity and higher concentration across the diversity and concentration metrics (all $p < .001$). ChatGPT 5.1 and Mistral Medium 3.1 showed a more mixed pattern: both had lower LexDiver and Hapax than Gemini 2.5 ($p < .001$), but higher Entropy and lower concentration on pTop10, Yule's K, and Simpson concentration (all $p < .05$ or stronger). These patterns show that model choice has a strong effect on the form of the generated data and that model comparisons depend on which aspect of diversity is being emphasized. In particular, a model may produce a more even word-frequency distribution while still generating fewer rare or one-off terms.

We repeated the generation settings used for the previous set of scatterplots to compare human responses with LLM responses. We selected normalized entropy and Simpson concentration as representative metrics because they summarize complementary aspects of the full word-frequency distribution. Normalized entropy captures the evenness of word use across the observed vocabulary and is less narrowly focused than lexical diversity or hapax rate, which are more sensitive to corpus length and rare-word counts. Simpson concentration captures the extent to which responses are dominated by repeated terms and has a direct probabilistic interpretation as the likelihood that two randomly selected word tokens are the same type. We reversed the Simpson concentration metric, so that the most diverse responses gave the highest values for both metrics. The resulting scatterplots are given in Figures 5 to 7.

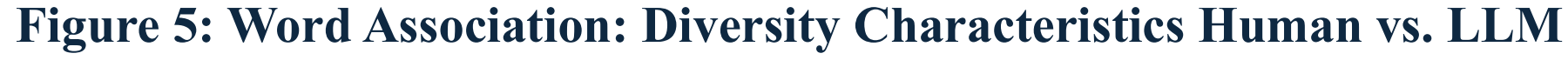

**Figure 5: Word Association: Diversity Characteristics Human vs. LLM**

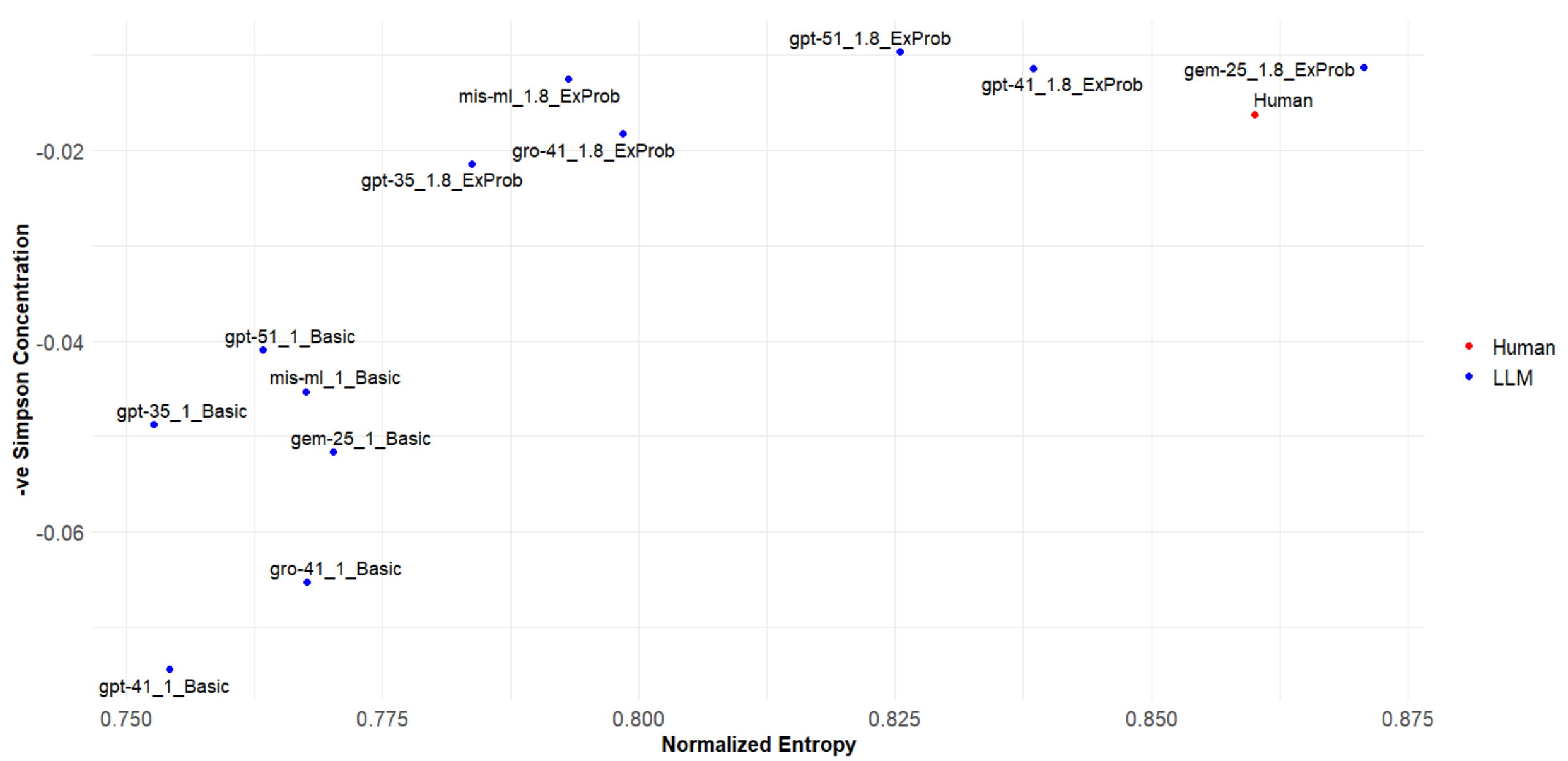

**Figure 6: Sentence Positive Diversity Characteristics Human vs. LLM**

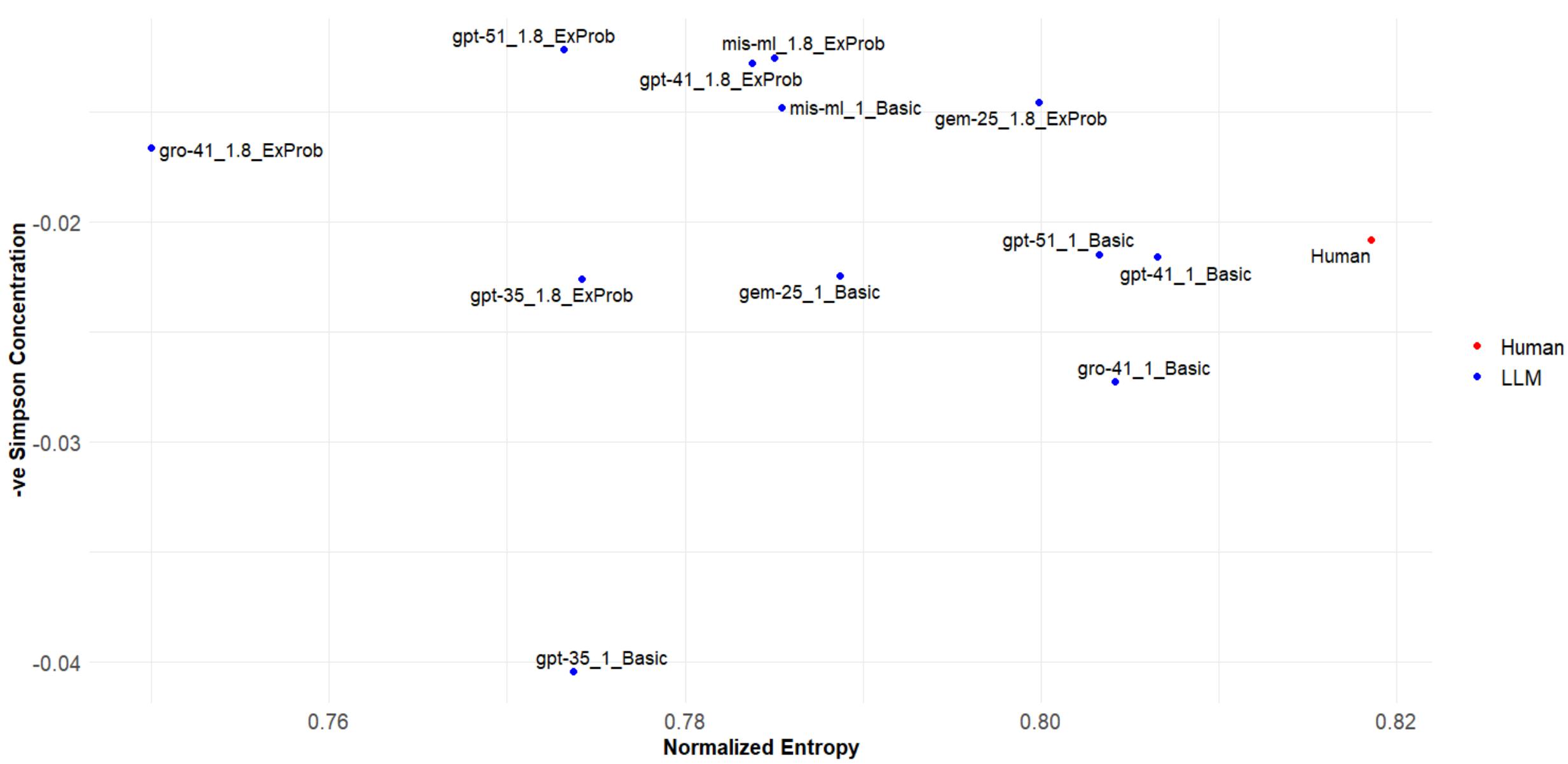

**Figure 7: Sentence Negative Diversity Characteristics Human vs. LLM**

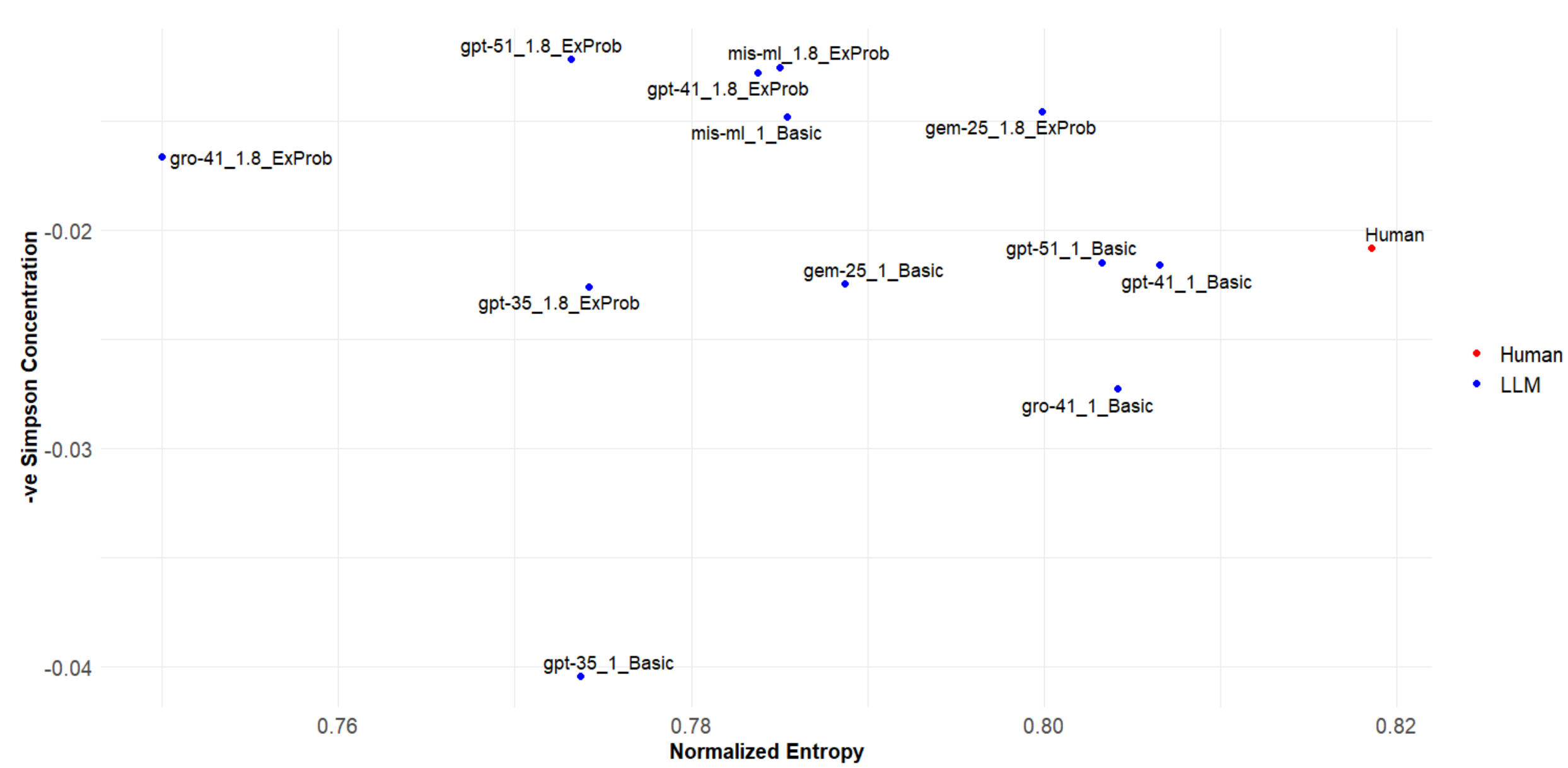

The human responses showed the highest level of normalized entropy for both sentence completion scenarios (negative and positive) and the second highest level of entropy for word association, with the Gemini 2.5 high-temperature ExtendedProb LLM condition having the first. The results are more mixed for reversed Simpson concentration. Human responses had average concentration for all three scenarios. This difference is likely because normalized entropy and Simpson concentration capture related but not identical properties of the word-frequency distribution. Normalized entropy is sensitive to the evenness of word use across the full vocabulary, whereas Simpson concentration is more strongly affected by the dominance of high-frequency repeated terms. Thus, the human responses may use a broad and relatively even vocabulary overall, while still sharing some common destination associations that produce moderate concentration.

### Case Study: Topic Modeling and Word Association

The preceding analyses examined how projective task, LLM model, prompting strategy, and temperature affected the distributional characteristics of LLM-generated responses, including response length, vocabulary breadth, diversity, and concentration. These analyses also identified LLM parameter settings that produced diversity characteristics similar to those of the human benchmark.

However, distributional similarity does not guarantee substantive similarity. LLM responses may resemble human responses in entropy or concentration while still emphasizing different topics, omitting important human associations, or introducing irrelevant content. Thus, to compare LLM and human responses we need to look more deeply at the topics of the responses. We do this using topic modeling. Topic models are used to identify latent themes in textual data and summarize the substantive content of large text corpora. They have been widely used in information systems and marketing research to extract consumer perceptions, product attributes, brand meanings, and customer needs from user-generated content such as online reviews, forums, and social media posts (Abramova et al., 2022; Netzer et al., 2012; Tirunillai & Tellis, 2014; Timoshenko & Hauser, 2019; Yang & Subramanyam, 2023; Zhang et al., 2025). In the present study, topic modeling allows us to move beyond distributional similarity and evaluate whether LLM-generated responses reproduce the substantive themes present in human projective responses.

We used the *stm* package in R to estimate structural topic models (Roberts et al., 2016; Roberts et al., 2019). Topic modeling represents a text corpus as a mixture of latent topics, where each topic is defined by a probability distribution over words and each document is represented by a probability distribution over topics. The model estimates these latent distributions so that the observed pattern of word use across documents is likely under the fitted topic structure. A useful feature of structural topic modeling is that document-level covariates can be incorporated into the model to explain variation in topic prevalence or topic content. In the present analysis, we include data from all cities and use city as a topic-prevalence covariate because different tourism destinations are likely to have different topic distributions, with some themes shared across cities and others more specific to a particular destination. For example, all cities may share broad tourism topics such as food, entertainment, safety, or crowds, but the prevalence of these topics may differ across Las Vegas, Los Angeles, Nashville, New Orleans, and New York City.

The number of topics chosen in a topic model is typically guided by a combination of model diagnostics and substantive interpretation. This decision can be based on model fit metrics, as well as by substantive researcher evaluation of the face validity of the topics. The *stm* package provides several model-fit metrics. Exclusivity measures the extent to which high-probability words are distinctive to a given topic rather than shared across many topics (Roberts et al., 2019). Held-out likelihood evaluates how well the model predicts unseen text and is commonly used as a measure of predictive fit in topic models (Wallach et al., 2009; Roberts et al., 2019). Residual diagnostics assess whether systematic word co-occurrence remains unexplained by the model, with lower residual dispersion indicating better fit (Taddy, 2012; Roberts et al., 2019). Semantic coherence measures whether the highest-probability words within a topic tend to co-occur in the same documents and is often used as a proxy for topic interpretability (Mimno et al., 2011; Roberts et al., 2019).

For parsimony, we focused on the word association task. First, we analyzed the n = 173 human responses, analyzing both individual words and bigram combinations of two words. Including bigrams allows for more flexible concepts; for example, “Mardis Gras” is a key concept for New Orleans. The model fit metrics for these data are plotted in Figure 8. We picked a 9-topic solution as this had the highest value for held-out likelihood, local maxima for exclusivity and semantic coherence, and a moderate value for residuals. It also had a good mixture of general topics and topics focused on one of the analysis cities.

**Figure 8: Word Association Human Topic Model Selection**

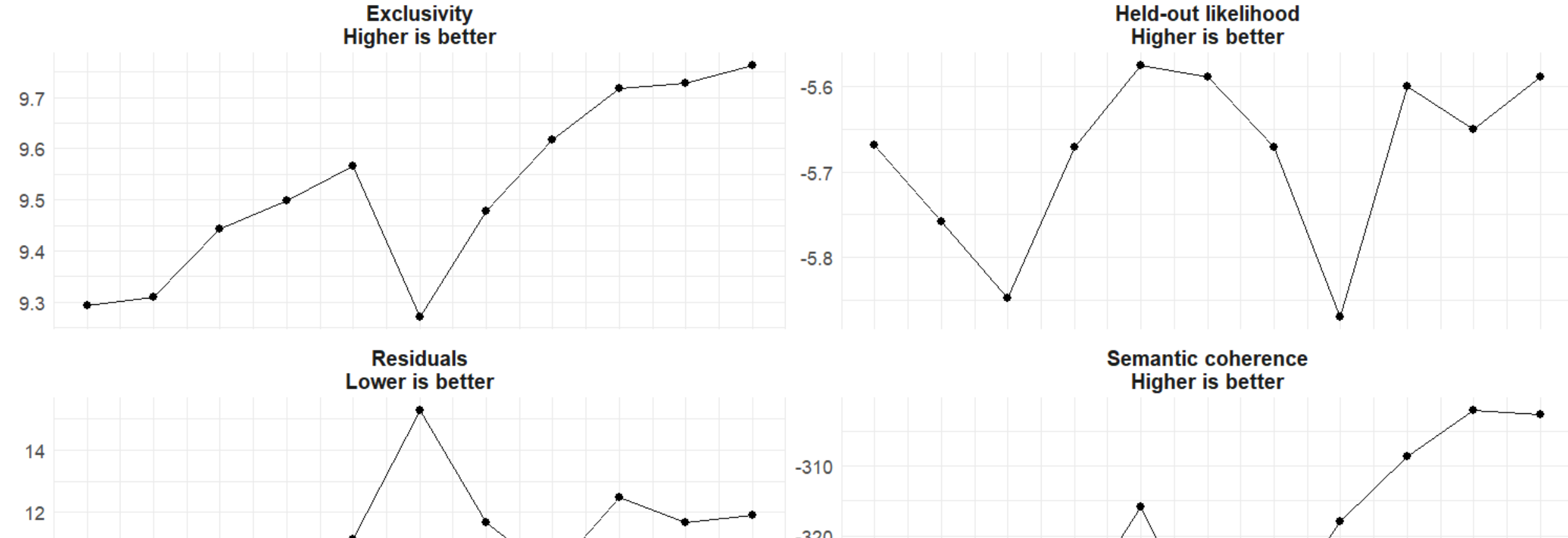

The nine topics are given in Table 7. The *stm* packages gives multiple "top k" lists of topic words. The "Highest Prob" terms are the highest value of the probabilities directly from the word-topic distribution. Score gives a heuristic measure that considers overall corpus word frequency distributions and the exclusivity of words in these distributions. We used these lists of words to name each of the topics, creating names that summarized the content in the top term lists. We can see a mixture of multi-city topics, such as "Expensive & Crowded" and topics focused on a single city, for example, "Gambling & Bright Lights", which was focused on Las Vegas.

**Table 7: Word Association Human Topic Modeling Summary**

| Tp | Top Words and Bigrams | Name |
|---|---|---|
| 1 | **Highest Prob:** food, good, street, gras, mardi, mardi_gras, life, bourbon, night, good_food<br>**Score:** good, life, gras, mardi, mardi_gras, night, food, street, night_life, good_food | Street Food & Entertainment |
| 2 | **Highest Prob:** casinos, hot, celebrities, traffic, beautiful, desert, history, dry, gambling, gambling_casinos<br>**Score:** celebrities, hot, casinos, beautiful, traffic, desert, history, dry, art, day | Casinos & Celebrities |
| 3 | **Highest Prob:** fun, hollywood, downtown, walk, culture, crowded, expensive, city, awesome, bars<br>**Score:** fun, hollywood, walk, awesome, downtown, walk_fame, super, culture, friends, crowded | Walking & Crowds |
| 4 | **Highest Prob:** big, busy, gambling, city, bright, food, money, lights, party, apple<br>**Score:** big, busy, gambling, walking, apple, vegas, big_apple, bright, party, lights | Gambling & Bright Lights |
| 5 | **Highest Prob:** people, lots, famous, homeless, famous_people, casinos, unique, many, homeless_people, lot<br>**Score:** people, lots, famous, famous_people, homeless, bustling, homeless_people, lots_people, many, unique | People of the City |
| 6 | **Highest Prob:** music, country, country_music, jazz, nightlife, city, concerts, jazz_music, nashville, atmosphere<br>**Score:** music, country, country_music, nightlife, jazz, atmosphere, jazz_music, club, good_music, concerts | Music & Concerts |
| 7 | **Highest Prob:** loud, casino, exciting, partying, wild, beignets, voodoo, smelly, active, sphere<br>**Score:** loud, partying, casino, exciting, wild, beignets, voodoo, clubbing, smelly, sphere | Partying & Casinos |
| 8 | **Highest Prob:** expensive, crowded, shopping, bars, southern, rich, sunny, movies, beach, beaches<br>**Score:** crowded, expensive, shopping, bars, confusing, southern, rich, streets, movies, sunny | Expensive & Crowded |
| 9 | **Highest Prob:** dirty, never, big_city, fame, cold, humid, city_never, never_sleeps, sleeps, tourist<br>**Score:** dirty, never, city_never, never_sleeps, sleeps, fame, big_city, tourist, cold, humid | Big City Issues |

We repeated the topic modeling process for the Gemini 2.5 LLM, with ExtendedProp prompt and a temperature of 1.8. We picked this combination of the settings, as it had, per Figure 5, the closest diversity characteristics to human responses. The model diagnostics for these data are plotted in Figure 9.

**Figure 9: Word Association LLM (Gemini2.5-ExtendedPropT=1.8) Topic Model Selection**

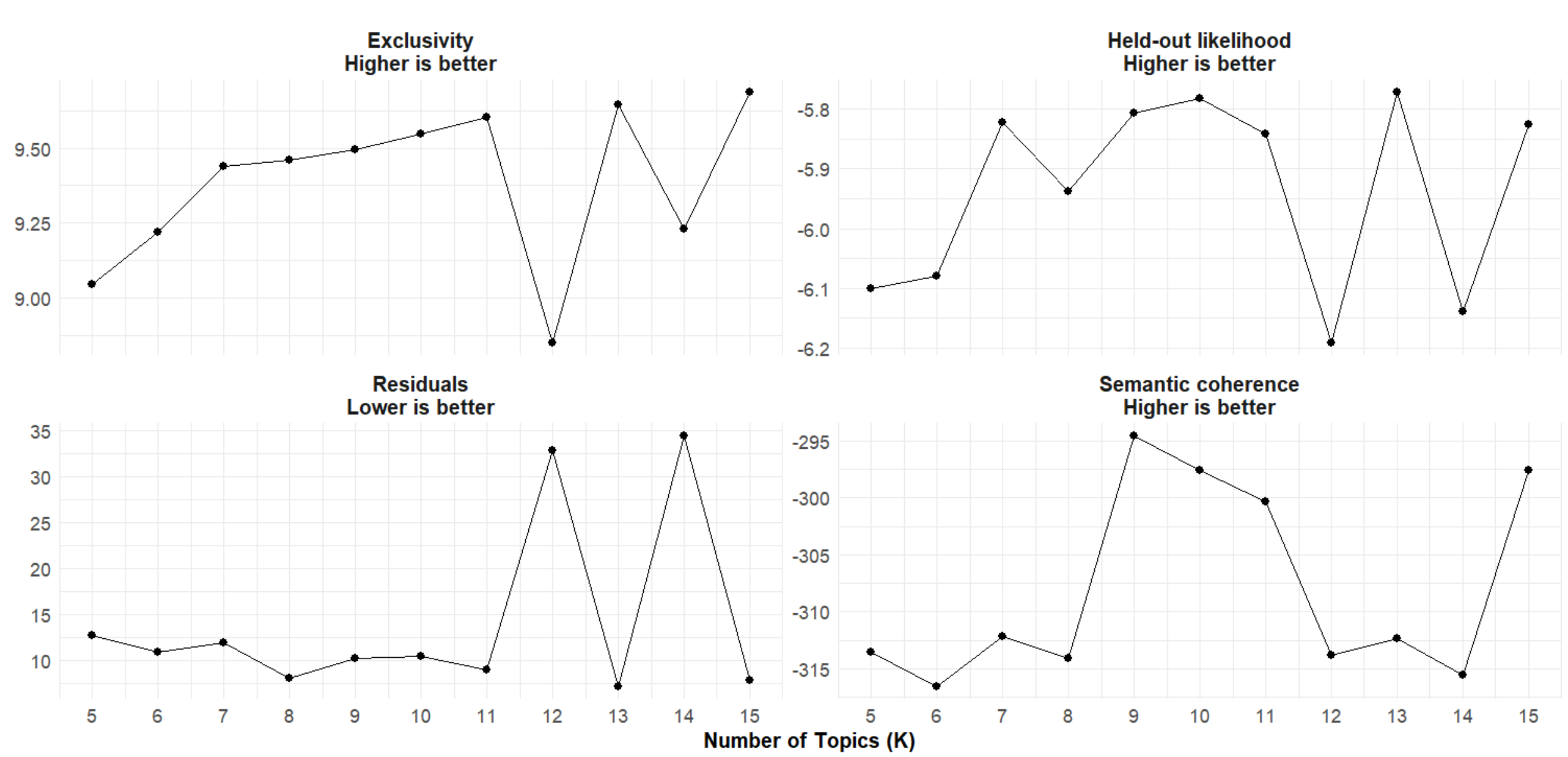

We chose the 9-topic solution, as it has the highest semantic coherence, strong values of the other metrics, and matches the number of topics for the human solution, to allow ease of comparison.

The topic details are output in Table 8. There is a high concordance between the topics generated from the human responses and from the LLM responses. Topics such as Hollywood crowds and drinking and partying in New Orleans and Nashville are shared between the topics. However, the LLM responses and human responses focus on different topics. For example, for Nashville, LLM responses have topics focused on Nashville hot chicken and Bachelorette parties that are less apparent in the human responses. Investigating this further, some of the individual responses included the word *casino* or *casinos* but there were not enough mentions for the model to include an entire casino topic.

An interesting insight from Table 8 is that the word diversity in the LLM solution seems to be generated by complex compound adjectives, for example *bachelorettepacked*, *walletemptying*, and *gumboscented*, rather than true diversity of topics.

**Table 8: Word Association LLM (Gemini2.5-ExtendedPropT=1.8) Topic Modeling Summary**

| Tp | Top Words and Bigrams | Name |
|---|---|---|
| 1 | **Highest Prob:** hot, chicken, hot_chicken, heaven, paradise, spicy, chicken_heaven, sunshine, food, beach<br>**Score:** chicken, hot_chicken, hot, chicken_heaven, paradise, spicy, sunshine, heaven, food, beach | Chicken Heaven |
| 2 | **Highest Prob:** bachelorette, central, party, gridlock, party_central, city, bachelorette_party, southern, little, bootscootin<br>**Score:** bachelorette, central, party, bachelorette_party, party_central, city, gridlock, bonanza, southern, bachelorette_central | Bachelorette Party Central |
| 3 | **Highest Prob:** traffic, starstudded, pricey, fun, iconic, wild, playground, sprawling, totally, overwhelmingly<br>**Score:** traffic, starstudded, playground, iconic, traffic_nightmare, wild, totally, nightmare, pricey, walletemptying | Big City Ups & Downs |
| 4 | **Highest Prob:** chaos, honkytonk, heaven, vibrant, money, chaotic, honkytonk_heaven, boozy, money_pit, pit<br>**Score:** chaos, honkytonk, heaven, honkytonk_heaven, vibrant, honky, honky_tonk, tonk, money_pit, pit | Honkytonk Chaos |
| 5 | **Highest Prob:** gritty, charm, hollywood, gritty_charm, glamorous, glam, glamour, yet, charmingly, glitz<br>**Score:** gritty, charm, gritty_charm, hollywood, glam, glamorous, glamour, hollywood_glam, charmingly, glitz | Charm, Grit, & Glamor |
| 6 | **Highest Prob:** overthetop, jazzsoaked, expensive, glittering, walletdraining, decadent, haunted, jazzfilled, mouthwatering, gridlocked<br>**Score:** overthetop, jazzsoaked, expensive, glittering, walletdraining, decadent, jazzfilled, haunted, fabulous, fashionforward | Expensive & Glittering |
| 7 | **Highest Prob:** nonstop, overload, endless, neon, energy, sensory, sensory_overload, dreams, sparkling, jungle<br>**Score:** nonstop, overload, sensory_overload, sensory, endless, neon, movie, jungle, energy, concrete | Sensory Overload |
| 8 | **Highest Prob:** streets, sticky, boujee, spicysweet, glitzy, mayhem, sticky_streets, bachelorettepacked, budget, neondrenched<br>**Score:** streets, boujee, sticky_streets, spicysweet, sticky, gumboscented, bachelorettepacked, mayhem, budget, glitzy | City Streets |
| 9 | **Highest Prob:** electric, historic, sweet, sweaty, charming, grimy, soulful, dream, slightly, trendy<br>**Score:** electric, dream, historic, charming, sweaty, grimy, soulful, hazy, sweet, slightly | City Charms & Descriptives |

To further investigate this issue, we created lists of "top terms" for the human responses and for the two LLMs with the closest diversity characteristics to the human responses, allowing for longer phrases including bigrams and trigrams. We first utilized the *quanteda* package in R to create candidate bigrams and trigrams. As in the previous analysis, we removed punctuation, converted all text to lowercase, and tokenized the responses into individual words[2]. We then identified frequently occurring multiword phrases and compounded them into single tokens, beginning with trigrams that appeared more than once. For example, "empire state building" was converted to empire_state_building. We then applied the same procedure to frequent bigrams. Because trigrams were compounded before bigrams, longer phrases were preserved before shorter overlapping phrases could be formed. In some cases, the process also produced longer compound tokens when overlapping trigrams and bigrams were merged across adjacent words. For example, "happens_in_vegas_stays_in_vegas" was created for human responses.

We then created a measure of lift, for each included word or phrase. The lift measure calculates how frequent a term is for a given city relative to its frequency in the remainder of the corpus (the other four cities). This measure is designed to find unique city descriptors rather than general descriptors and thus help elicit key city insights. For term $j$ in city $c$, the lift measure is calculated as follows:

$$textLift_{cj} = \frac{log_2(P(j|c))}{log_2(P(j|\text{not } c))} \text{ where } P(j|c) = \frac{n_{cj}+\alpha}{N_c+\alpha} \text{ and } P(j|\text{not } c) = \frac{n_{-c,j}+\alpha}{N_{-c}+\alpha}, \quad (1)$$

where $n_{cj}$ is the count of term $j$ in city $c$, $N_c$ is the total number of terms in city $c$, $n_{-c,j}$ is the count of term $j$ in all other cities, $N_{-c}$ is the total number of terms in all other cities. The smoothing constant, set at $\alpha = 0.5$, prevents indeterminate values when a term is only present in one city. The top 10 terms for each city are given for human respondents in Figure 10, for Gemini2.5 with ExtendedProp prompt and Temperature=1.8 in Figure 11, and for ChatGPT4.1 with ExtendedProp prompt and Temperature=1.8 in Figure 12.

[2] We also produced versions of this analysis for stemmed terms. However, after looking at initial results, for this analysis, we felt that unstemmed terms gave clearer interpretation.

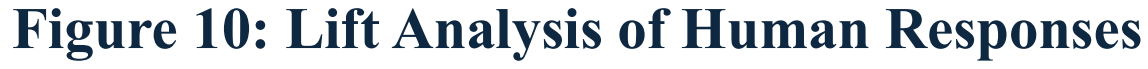

# Figure 10: Lift Analysis of Human Responses

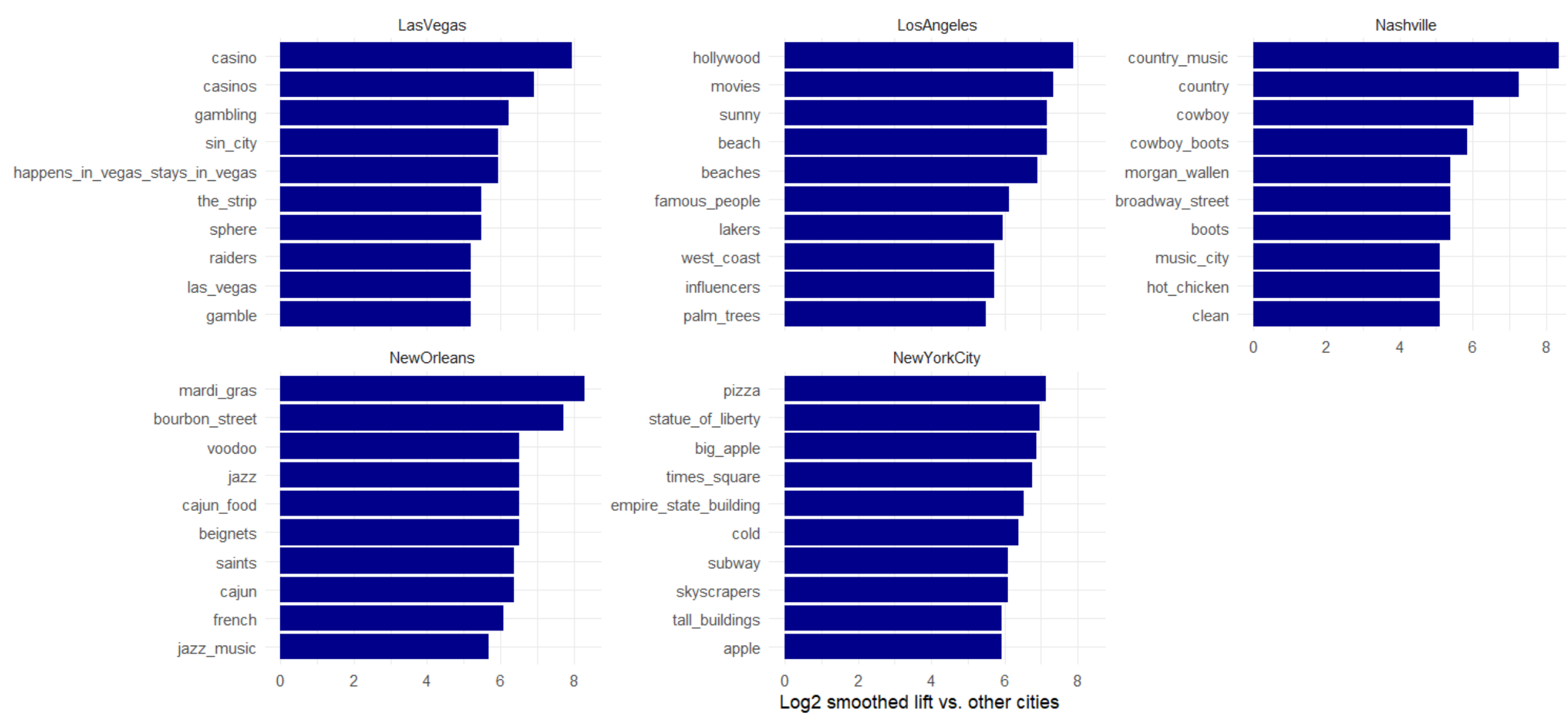

# Figure 11: Lift Analysis of Gemini2.5-ExtendedPropT=1.8 Responses

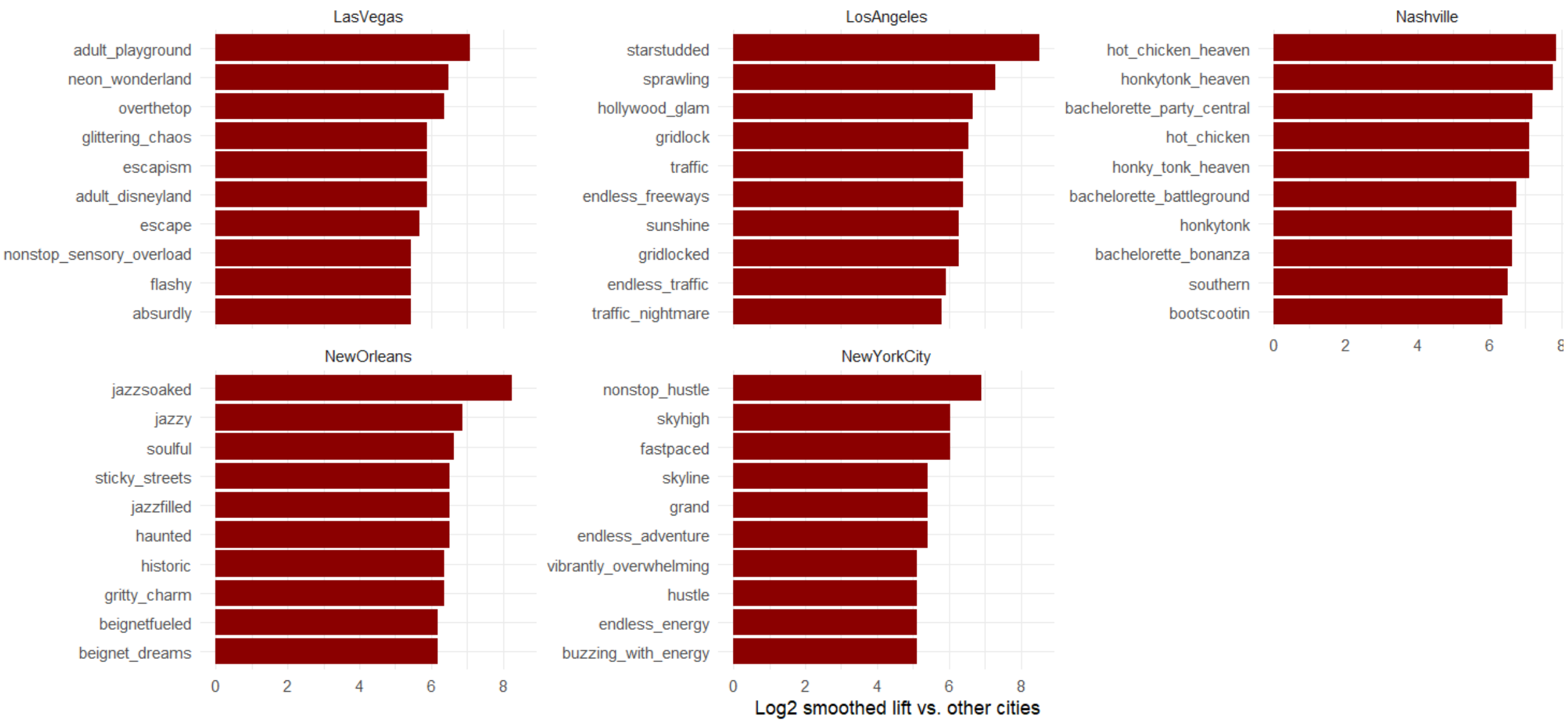

**Figure 12: Lift Analysis of ChatGPT4.1-ExtendedPropT=1.8 Responses**

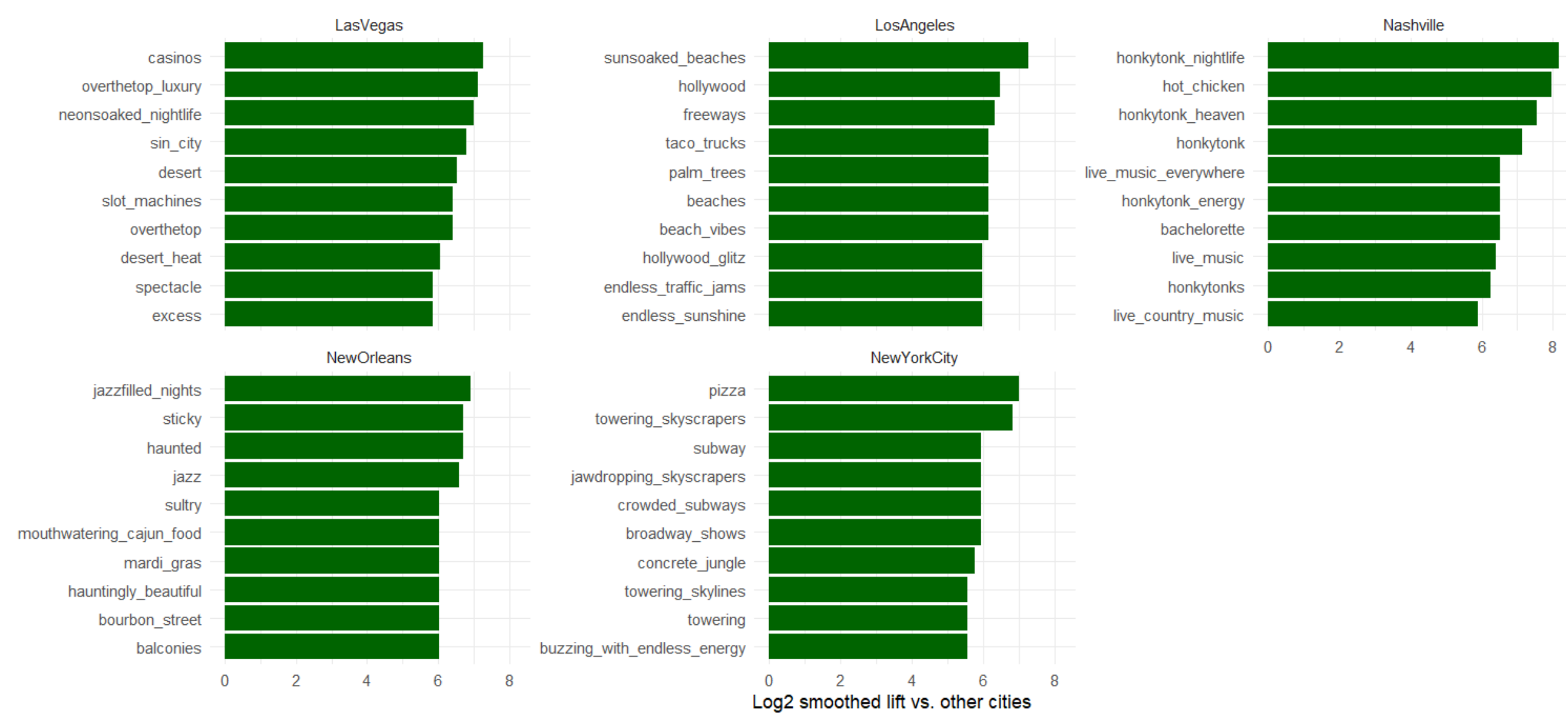

The LLM results covered many of the same terms as the human results, but with different emphases. For example, for Nashville, the top human responses focused on country music, cowboy culture, and hot chicken. The LLM results were similar, but the top terms for both LLMs included bachelorette parties (which was mentioned in the human responses but was not in the top 10 terms). As noted in the topic models, both the human and LLM responses have similar diversity characteristics, but the diversity is created in different ways. The Gemini-2.5 responses include complex compound adjectives, such as *beignetfueled*. Both of the tested LLM response sets include creative descriptive phrases, such as *nonstop_sensory_overload* for Gemini-2.5 and *mouthwatering_cajun_food* for ChatGPT4.1.

## DISCUSSION AND IMPLICATIONS

At the beginning of the research process, we posed four research questions. We analyze these questions in detail below given the results of the empirical work.

First, can LLMs generate synthetic consumer responses to projective techniques that resemble responses provided by human participants and give equivalent insights? The answer to this question is that by controlling prompting strategy and how LLMs sample from the distribution of responses via temperature settings, LLMs can generate realistic responses that have similar diversity characteristics and that cover similar topics to human responses. However, there are some stylistic differences between human responses and LLM responses, for example, with LLMs creating complex compound adjectives to describe places and experiences. These stylistic differences likely reflect differences in the response-generation process. Human respondents in projective tasks typically provide brief associations based on memory, experience, and familiar cultural images. LLMs, by contrast, generate text by sampling from patterns learned from large corpora, with outputs further shaped by prompt wording, temperature, and instruction tuning (Brown et al., 2020; Ouyang et al., 2022). This can lead LLMs to reproduce similar topical associations while expressing them in a more polished or stylized form. Higher-temperature sampling may also encourage less typical word combinations, increasing apparent diversity while sometimes producing more artificial phrasing (Holtzman et al., 2019). Therefore, the LLM responses may be substantively similar to human responses at the level of broad city associations, but stylistically different in how those associations are expressed.

Second, what metrics are appropriate for evaluating the similarity, diversity, and interpretive richness of LLM-generated projective data? We utilized a range of different metrics to compare human responses with LLM responses. We calculated a set of basic text characteristics, which include the average response word count, the average word length, the proportion of stop words, and the total vocabulary size across responses. In the experiment, these metrics showed a high ability to distinguish the response characteristics for different LLMs and LLM prompting and temperature settings, and for regression models with the experimental settings as IVs and the metrics as DVs, the $R^2$ of the regression models ranged from 0.538 (overall vocabulary size) to 0.896 (proportion of stopwords). We also calculated a range of text concentration and diversity metrics, each testing the breadth of responses using slightly different mechanics. These metrics are important in the context of using projective techniques for idea generation, as a narrow range of repeated responses would limit the insights available from the data. Again, these metrics showed a high ability to differentiate between experimental settings, with the $R^2$ of the regression models with these metrics as DVs ranging from 0.498 (Hapax) to 0.698 (entropy). We also used topic modeling and top 10 term analyses to explore the range of content contained in the human and LLM responses. While these methods are quality metric per-se, understanding the differences in topic structure and phrase structure for the responses allowed us insight into some of the insights into the different generation processes behind human and LLM responses.

Third, how do prompting strategies, seed examples, and model parameters affect the quality of generated responses? The results show that LLM-generated projective data are highly sensitive to researcher-controlled design choices. This finding connects directly to the argument developed in the introduction: synthetic consumer data generated by LLMs are not a fixed output of the model alone, but the result of an interaction among the model, prompt design, examples, and sampling parameters. Consistent with prompt-engineering research showing that LLM outputs depend substantially on task framing, examples, and instructions (White et al., 2023; Sahoo et al., 2024; Schulhoff et al., 2024), more detailed prompts with explicit instructions for creativity and broader response sampling produced more diverse outputs. Similarly, the finding that seed human examples narrowed the range of responses and tied the outputs more closely to the human benchmark is consistent with research on few-shot prompting, where examples guide model behavior (Brown et al., 2020; Min et al., 2022). Finally, as expected from the role of temperature in probabilistic text generation, higher temperature settings produced more diverse responses by increasing the likelihood of less typical word choices.

Fourth, do different LLMs, and different generations of LLMs, vary in their ability to generate projective consumer data? We found significant differences in responses between the tested LLMs. For example, ChatGPT5.1 gave more verbose, diverse responses than older versions of ChatGPT. Overall, Gemini and ChatGPT4.1 on high temperature settings and with extended prompts gave good diversity results for the word association tasks, though as described earlier, the generated diversity was a little different from the human response diversity. Our results concur with recent LLM evaluation research, which emphasizes that model performance varies across tasks, metrics, and evaluation settings, and that no single model is likely to dominate across all use cases (Liang et al., 2022; Shnitzer et al., 2023).

**Overall Managerial Implications**

A key overall question is can we use LLMs to generate synthetic data for creative marketing tasks? Overall, the answer is not a simple yes or no. The results suggest that LLMs can be useful for generating synthetic data for creative marketing tasks, but they should be used with care. LLMs can generate useful ideas, themes, and associations, but they should not be treated as a replacement for human respondents when the goal is to estimate how common a belief or perception is. For example, LLM data should not be used to conclude that "30% of consumers think casinos are central to the Las Vegas experience."

Where LLMs appear most useful is in helping researchers explore the range of possible consumer associations. For projective techniques, where the goal is often to uncover ideas, meanings, emotions, and unexpected connections, LLMs can produce responses that overlap substantially with human responses and do so at much lower cost. This makes them valuable as a supplement to human data collection, especially when budgets or timelines are limited.

When data collection is expensive LLMs could be used to supplement human data collection to gain additional ideas and insights, either through separate and complementary data collection or by using human responses to guide LLM responses through "few-shot" prompting scenarios. In addition, LLM responses could be used as a cheap method of piloting qualitative and mixed methods research projects before moving to humans for more accurate, grounded insights.

As a note of caution, LLM responses should not be treated as equivalent to human insight. The models often produce more polished, elaborate, and stylized language than human respondents. Sometimes this creates useful creative variation, but it can also make the responses feel less grounded in lived experience. Therefore, LLMs are best viewed as a low-cost tool for expanding and piloting consumer insight work, rather than as a full substitute for human respondents in creative or qualitative marketing research.

## LIMITATIONS AND FUTURE WORK

We analyzed LLM responses to two different projective techniques, word association and sentence correction. These methods extract relatively short succinct human responses. Projective techniques include more complex creation and construction techniques (e.g., Lilienfeld et al., 2000; Zayer et al. 2024), for example, having respondents imagine a brand or retail consumption scenario and writing a creative essay or play (Hindley & Font, 2018). Such creative endeavors are within the capabilities of generative AI (Marco et al., 2025) and the experiments in this paper could be extended to more complex projective-technique scenarios involving storytelling, creative writing, or scenario construction.

Projective techniques can also involve the creation of images. One of the original projective techniques used in psychology was the draw-a-person test (Machover, 1949), where respondents draw male and female figures and then these drawings are interpreted. In a consumer research context, Gamradt (1995) ran a projective data gathering exercise where Jamaican children drew representations of tourists and tourism scenes. Given this use of images to extract deeper insight in projective techniques and the increasing power of LLMs to create images (Bansal et al., 2024) the research in this paper could be expanded to image generation procedures.

The experiments in this dataset involved replicating data gathered from college students from a single university. Because of this and the fact that we wanted a simple prompting strategy that worked across multiple LLMs, we didn't add additional demographic variables (for example, age, ethnic group, country region, profession, etc….) or create complex personas. Targeting different segments of consumers is key to marketing strategy, so there is scope to expand the work to include more complex demographic targeting and persona generation (e.g., Ge et al., 2024; Kang et al., 2024) and more complex, targeted prompts.

This study evaluates LLM-generated synthetic data primarily by analyzing the textual responses from the LLMs. In reality, the text generation process is part of a complex system involving specifying the prompts, analyzing the prompts, and utilizing the responses to help develop marketing strategy. All these tasks require human-AI interaction. Future research could also examine LLM-generated synthetic data as a human-AI interaction process. This would connect the study to the IS tradition of human-computer interaction, which examines how users interact with technologies, tasks, and information in managerial contexts (Zhang et al., 2002). It would also connect to IS research on agentic information systems and human-AI collaboration, where the value of AI depends not only on model output and performance but also on how humans work with AI systems (Baird & Maruping, 2021; Fügener et al., 2022)